%% file: arXiv.tex
\documentclass[10pt,onecolumn,letterpaper]{article}

\usepackage[top=1.5in,bottom=1.5in,left=1.2in,right=1.2in,columnsep=0.35in]{geometry}

\usepackage{indentfirst}
\usepackage{xcolor}
\usepackage{hyperref}
\usepackage{enumitem}
\usepackage{setspace}
\usepackage{algorithm}
\usepackage{algorithmic}
\usepackage{amsmath}
\usepackage{amsthm}
\usepackage{amssymb}
\usepackage{natbib}
\usepackage{stmaryrd}
\usepackage{hyperref}
\usepackage{tabularx}
\usepackage{booktabs} 
\usepackage{color}
\usepackage{graphicx}
\usepackage{caption}
\usepackage{subfigure}
\usepackage{pifont}

\usepackage{authblk}

\newtheorem{lemma}{Lemma}[section]
\newtheorem{theorem}{Theorem}[section]
\newtheorem{corollary}{Corollary}[section]

\newtheorem{assumption}{Assumption}        
\theoremstyle{definition}
\newtheorem{definition}{Definition}
\newtheorem{remark}{Remark}[section]

\author[1]{Wenjie Li}
\author[1]{Qifan Song}
\author[2]{Jean Honorio}
\affil[1]{Department of Statistics, Purdue University}
\affil[2]{Department of Computer Science, Purdue University}

\date{}

\begin{document}

\title{Personalized Federated $\mathcal{X}$-armed Bandit}
\maketitle

\begin{abstract}
In this work, we study the personalized federated $\mathcal{X}$-armed bandit problem, where the heterogeneous local objectives of the clients are optimized simultaneously in the federated learning paradigm. We propose the \texttt{PF-PNE} algorithm with a unique double elimination strategy, which safely eliminates the non-optimal regions while encouraging federated collaboration through biased but effective evaluations of the local objectives. The proposed  \texttt{PF-PNE} algorithm is able to optimize local objectives with arbitrary levels of heterogeneity, and its limited communications protects the confidentiality of the client-wise reward data. Our theoretical analysis shows the benefit of the proposed algorithm over single-client algorithms. Experimentally, \texttt{PF-PNE} outperforms multiple baselines on both synthetic and real life datasets.
\end{abstract}

\input{section_manager}

\end{document}

%% file: section_manager.tex
\newenvironment{list1}{
  \begin{list}{$\bullet$}{%
      \setlength{\itemsep}{3pt}
      \setlength{\parsep}{5pt} \setlength{\parskip}{0in}
      \setlength{\topsep}{5pt} \setlength{\partopsep}{0in}
      \setlength{\leftmargin}{15pt}}}{\end{list}}

\newenvironment{list2}{
  \begin{list}{$\bullet$}{  
      \setlength{\itemsep}{5pt}
      \setlength{\parsep}{5pt} \setlength{\parskip}{0in}
      \setlength{\topsep}{5pt} \setlength{\partopsep}{0in}
      \setlength{\leftmargin}{30pt}}}{\end{list}}

\input{Sections/01.Introduction}

\input{Sections/02.Prelim}
\input{Sections/03.Algorithm}
% \clearpage
\input{Sections/04.Experiments}

\input{Sections/05.Conclusions}

\clearpage

\bibliographystyle{plainnat}
\bibliography{my_ref}

\clearpage
\onecolumn
\appendix

\centerline{\textbf{\LARGE Appendix to ``Personalized Federated $\mathcal{X}$-armed Bandit" }}

\input{Sections/Appendix01}

\input{Sections/Appendix02}
\input{Sections/Appendix03}

%% file: Sections/01.Introduction.tex
\section{Introduction}
\label{sec: introduction}
Federated bandit is a novel research area that combines sequential decision-making with federated learning, addressing data heterogeneity and privacy protection concerns for trustworthy machine learning \citep{mcmahan2017communication, shi2021federateda}. Unlike traditional bandit models that focus solely on the exploration-exploitation tradeoff, federated bandit also considers the implications of modern data privacy concerns. Federated learning involves data from non-i.i.d. distributions, making collaborations between clients essential for accurate inferences for the global model. However, due to the concerns of communication cost and user privacy, these collaborations must be limited, and direct local data transmission is avoided. To make accurate decisions, clients must coordinate their exploration and exploitation, utilizing minimal communication among them.

Most existing federated bandit research has mainly focused on finite arms (i.e., multi-armed bandit) or linear contextual bandits, where the expected reward is a linear function of the chosen contextual vector \citep{shi2021federateda, shi2021federatedb, huang2021federated, dubey2020differentially}. Some recent works on neural bandits have extended the results to nonlinear reward functions \citet{zhang2020neural, dai2023federated}. However, more complicated problems such as dynamic pricing and hyper-parameter optimization require solutions for domains with infinite or even uncountable cardinality, posing challenges to the current federated bandit algorithms' applicability in real-world scenarios. For example, when deploying base stations for different locations, several hyper-parameters need to be tuned for the best performance of the base stations. The hyper-parameters are often chosen from a fixed domain,  e.g., a hypercube in $\mathbb{R}^d$. The best set of hyper-parameters for different locations could be different, but the performance of a fixed set of hyper-parameters should be similar for locations that are close to each other, thus encouraging federated learning.

Several kernelized bandit algorithms are proposed to address such problems with nonlinear rewards and infinite arm domains \citep{chowdhury2017kernelized, li2022communication}. However, these works are based on very different assumptions from ours and have relatively high computational costs. The only work closely related to our research is \citet{Li2022Federated}. However, they only consider optimizing the cumulative regret on the \textit{global objective}, which refers to the average of all the client-wise local objectives and thus the best point ``on average". In our paper, we aim to optimize all the local objectives at the same time so that each client locates its own optimum. This is much more challenging but beneficial to real applications. We compare our work with some of the existing works in Table \ref{tab: regret_compare}.

\newcommand{\cmark}{\ding{51}}%
\newcommand{\xmark}{\ding{55}}%
\begin{table*}

\caption{
% \footnotesize 
Comparison of the average regret upper bounds, the communication cost for sufficiently large $T$ and the other properties. \textbf{Columns}: ``Commun. rounds'' refer to the number of communication rounds.
``Personalized" refers to whether the local objectives or the global objectives are optimized. 
\textbf{Rows:} 
% 
% \texttt{HOO} is a representative single-client $\mathcal{X}$-armed bandit algorithm. \texttt{BLiN} is a batched-$\mathcal{X}$-armed bandit algorithm. 
% 
\texttt{Centralized} results are adapted from the single-client $\mathcal{X}$-armed bandit algorithms such as \texttt{HOO} \citep{bubeck2011X} and \texttt{HCT} \citep{azar2014online} by assuming that the server makes all the decisions with access to all client-wise information. \texttt{Fed-PNE} is a federated $\mathcal{X}$-armed bandit algorithm that optimizes the global regret and is thus not personalized. Therefore, the comparisons with these two algorithms are not completely fair.
\textbf{Notations:} $M$ denotes the total number of clients; $T$ denotes the time horizon; for simplicity of comparison, we assume that all objectives $f_1, f_2, \cdots, f_M, \overline{f}$ share the same the near-optimality dimension denoted by $d$ (in Definition \ref{definition: near-optimality dimension}).; $d_{\text{new}} \leq d$ is the optimality-difference dimension (in Definition \ref{definition: optimality-difference dimension}). }
\label{tab: regret_compare}
% \vspace{5mm}
\centering
\begin{tabular}{lllc }
\hline Bandit algorithms & Average Regret  & Commun. rounds & Personalized\\
\hline  
%%%%%%%%%%%%%
\texttt{HOO} (\citet{bubeck2011X})
 & $\widetilde{\mathcal{O}}\left( T^{\frac{d+1}{d+2}}\right)$ & N.A. & \cmark\\
 
\texttt{BLiN} (\citet{Feng2021Lipschitz})
 & $ \widetilde{\mathcal{O}} \left(  T^{\frac{d+1}{d+2}} \right)$ & N.A. & \cmark \\

$\texttt{Centralized}^*$
 & $\widetilde{\mathcal{O}} \left( M^{-\frac{1}{d+2}}  T^{\frac{d+1}{d+2}} \right)$ & ${\mathcal{O}}(MT)$ & \xmark \\

$\texttt{Fed-PNE}^*$ (\citet{Li2022Federated}) & $\widetilde{\mathcal{O}} \left( M^{-\frac{1}{d+2}}  T^{\frac{d+1}{d+2}} \right)$ & $\widetilde{\mathcal{O}}(M \log T)$ & \xmark \\
\hline 

\texttt{PF-PNE}  (this work) & $\widetilde{\mathcal{O}} \left( M^{-\frac{1}{d+2}}  T^{\frac{d+1}{d+2}}  + T^\frac{d_{\text{mew}}  + 1}{d_{\text{new}}  + 2} \right) $ & $\widetilde{\mathcal{O}}(M \log T)$ & \cmark \\
\hline 
% \vspace{-30pt}
\end{tabular}
\end{table*}

We highlight our major contributions as follows. 
\begin{list1}
    \item \textbf{Personalized federated $\mathcal{X}$-armed bandit.} We propose the personalized federated  $\mathcal{X}$-armed bandit problem, where different clients optimize their own local objectives defined on a domain $\mathcal{X}$. The new problem is much more challenging than prior research due to the heterogeneity of the local objectives and the limited communication.

    \item \textbf{\texttt{PF-PNE} algorithm and double elimination.} We propose the first algorithm, \texttt{PF-PNE} to solve the personalized federated $\mathcal{X}$-armed bandit problem. The algorithm incorporates a novel double elimination strategy which guarantees that the non-optimal regions are only eliminated after thorough checks. The first round of elimination removes potential non-optimal regions and the second round of elimination uses biased evaluations, obtained from sever-client communications, to avoid redundant evaluations and unnecessary costs.

    \item  \textbf{Theoretical analysis and empirical evidence.}  Theoretically, we prove that the proposed algorithm enjoys a $\widetilde{\mathcal{O}} \left( M^\frac{d_{\min}  + 1}{d_{\min}  + 2} T^\frac{d_{\min}  + 1}{d_{\min}  + 2} + M T^\frac{d_{\text{new}}  + 1}{d_{\text{new}}  + 2}\right)$ regret bound (where ${d}_{\min}$ and $d_{\text{new}}$ are defined in Definitions \ref{definition: near-optimality dimension} and \ref{definition: optimality-difference dimension}). When the local objectives are similar, $d_{\text{new}}$ is very small, yielding smaller average regret than single client algorithms. Moreover, the algorithm only requires limited communications in total, which greatly protects user data confidentiality. Empirically, we provide evidence to support our theoretical claims on both synthetic objectives and real-life datasets. \texttt{PF-PNE} outperforms existing centralized and federated bandit algorithm baselines.

\end{list1}

%% file: Sections/02.Prelim.tex
\section{Preliminaries}
\label{sec: prelim}

In this section, we discuss the concepts, notations and assumptions used in this paper, most of which follow those used in \citet{Li2022Federated}. For an integer $N \in \mathbb{N}$, $[N]$ is used to represent the set of positive integers no larger than $N$, i.e., $\{1, 2,\cdots, N\}$. For a set $\mathcal{A}$, $|\mathcal{A}|$ denotes the number of elements in $\mathcal{A}$. For a real number $a\in\mathbb{R}$, we use $\lceil {a}\rceil$ and $\lfloor {a}\rfloor$ to represent the smallest integer larger than $a$, and the largest integer smaller than $a$ respectively. 
Throughout this paper, we use the subscript notation to represent the client (local) side definitions, e.g., the local objective $f_m$ and the local near-optimality dimension $d_m$. We use the overline notation to represent the server (global) side definitions, e.g., the global objective $\overline{f}$ and the global near-optimality dimension $\overline{d}$.
In big-$\mathcal{O}$ notations, we use $\widetilde{\mathcal{O}}(\cdot)$ to hide the logarithmic terms, i.e., for two functions $a(n), b(n)$, $a(n) =\widetilde{\mathcal{O}}(b(n)) $ represents that $a(n)/b(n)\leq \log^{k}(n), \forall n >0$ for some $k>0$.

\subsection{Problem Setting}
We denote the available measurable space of arms as $\mathcal{X}$. In accordance with the practical applications, we formulate the problem setting as follows: we assume that in total $M\in \mathbb{N}$ clients want to collaboratively solve their problems, and thus $M$ \textit{local objectives} are available, denoted by $\{f_m\}_{m=1}^M$, all defined on the same space $\mathcal{X}$ and bounded by $[0, 1]$. These local objectives could be non-convex, non-differentiable and even non-continuous. With a limited budget $T$, Each client can only evaluate its own {local objective} once per round by choosing an arm $x_{m, t} \in \mathcal{X}$ at each round $t \in [T]$ and then observes a noisy feedback $r_{m,t} \in [0, 1]$ defined as $ r_{m, t}:= f_m(x_{m, t}) + \epsilon_{m,t}$, where $\epsilon_{m,t}$ is a zero-mean and bounded random noise independent from previous evaluations and other clients' evaluations.

Similar to the prior federated bandit works such as \citet{shi2021federateda, huang2021federated, Li2022Federated}, we assume that a central server exists and it is able to communicate with the clients in every round. To protect user privacy and confidentiality, the central server can only share summary statistics of the rewards (e.g., the empirical mean and variance) from different clients. The original rewards of each evaluation should be kept confidential. The clients are not allowed to communicate with each other and we assume that the server and all the clients are completely synchronized \citep{mcmahan2017communication, shi2021federateda}. Note that the number of clients $M$ could be very large and thus incurring very high communication costs when the clients choose to communicate with the server. Therefore, we need to take into consideration such costs in the algorithm design and the analysis. This work aims to design an algorithm, that adapts to the heterogeneity among local objectives, such that collaborative search helps when the local objectives are similar.

\subsection{Performance Measure}

In the setting of \citet{Li2022Federated}, the clients are required to jointly solve for the \textit{global maximizer}, i.e., the objective is to find the point $x$ that maximizes the \textit{global objective}, $ \overline{f}(x) := \frac{1}{M} \sum_{m=1}^M f_m (x)$. However, as we have mentioned in the examples in Section \ref{sec: introduction}, very often the global maximizer is not the best option for every client and the clients would want to maximize their own benefit by finding the maximizer of their local objectives. Therefore, instead of the \textit{global regret} defined in \citet{Li2022Federated} where the performance of the clients is measured on the global objective, we want to minimize the expectation of the \textit{local cumulative regret}, defined as follows
\begin{equation}
\nonumber
    R(T) = \sum_{t=1}^T \sum_{m=1}^M f_m^*  - \sum_{t=1}^T \sum_{m=1}^M f_m(x_{m,t})
\end{equation}
where $f_m^*$ denotes the optimal value of $f_m$ on $\mathcal{X}$ and similar notation is used for $\overline{f}$. In order to find their own optimum, the clients can only utilize the noisy evaluations $r_{m,t}$ of their own local objective functions $f_m$, and the information communicated with the central server. Moreover, it is expected that some assumptions on the similarity between the local objectives and the global objective are necessary so that the communications are useful. Otherwise, the local objectives could be completely different and collaboration among the clients would be meaningless. We will discuss our assumption in Section \ref{subsec: definitions_and_assumptions}.

\subsection{Hierarchical Partitioning}
Similar to the existing works on $\mathcal{X}$-armed bandit \citep[e.g.,][]{azar2014online, shang2019general, bartlett2019simple, Li2022Federated}, our algorithms rely on the recursively-defined hierarchical partitioning $\mathcal{P} := \{\mathcal{P}_{h,i}\}_{h,i}$ of the parameter space $\mathcal{X}$. The hierarchical partition discretizes the space $\mathcal{X}$ into several nodes on each layer by the following relationship:
\begin{equation}
\nonumber
    \mathcal{P}_{0, 1} := \mathcal{X}, \qquad \mathcal{P}_{h,i} := \bigcup_{j=0}^{k-1}  \mathcal{P}_{h+1, ki-j},
\end{equation}
where for every node $\mathcal{P}_{h,i}$ inside the partition, $h$ and $i$ represent its depth and index respectively. For each $h \geq 0, i > 0, \{\mathcal{P}_{h+1, ki-j}\}_{j=0}^{k-1}$ are disjoint children nodes of the node $\mathcal{P}_{h,i}$ and  $k$ is the number of children for one node. 
The union of all the nodes on each depth $h$ equals the parameter set $\mathcal{X}$. The partition is settled and shared with all the clients and the central server before the federated learning process because the partition is deterministic and contains no information of the reward evaluations.

\subsection{Assumptions}
\label{subsec: definitions_and_assumptions}

We first present the assumptions that are also observed in prior $\mathcal{X}$-armed bandit works \citep{bubeck2011X, azar2014online, Grill2015Blackbox, Li2022Federated}. 

\begin{assumption}
\label{assumption: dissimilarity}
\textbf{\upshape(Dissimilarity Function)}
        The space $\mathcal{X}$ is equipped with a dissimilarity function $\ell: \mathcal{X}^2 \mapsto \mathbb{R}$ such that $\ell(x, x')\geq 0, \forall (x, x') \in \mathcal{X}^2$ and $\ell(x, x) = 0$ 
\end{assumption}
We will assume that Assumption \ref{assumption: dissimilarity} is satisfied throughout this work. Given the dissimilarity function $\ell$, the diameter of a set $\mathcal{A} \subset \mathcal{X}$ is defined as $\operatorname{diam}\left(\mathcal{A}\right) = \sup_{x, y \in \mathcal{A}} \ell(x, y) $. The open ball of radius $r$ and with center $c$ is then defined as $\mathcal{B}(c, r) = \{x \in \mathcal{X}: \ell(x, c) \leq r\}$. We now introduce the local smoothness assumptions.

\begin{assumption}\textbf{\upshape (Local Smoothness) }
\label{assumption: local_smoothness}
We assume that there exist constants $\nu_1, \nu_2 > 0$, and $0< \rho < 1$ such that for all nodes $\mathcal{P}_{h,i}, \mathcal{P}_{h,j}\in \mathcal{P}$ on depth $h$,
\begin{list1}
    \item $\operatorname{diam}\left(\mathcal{P}_{h, i}\right) \leq \nu_{1} \rho^{h}$
    \item  $\exists x_{h, i}^{\circ} \in \mathcal{P}_{h, i}$ s.t. $\mathcal{B}_{h, i} {:=} \mathcal{B}\left(x_{h, i}^{\circ}, \nu_{2} \rho^{h}\right) \subset \mathcal{P}_{h, i}$
    \item $\mathcal{B}_{h, i} \cap \mathcal{B}_{h, j}=\emptyset$ for all $1 \leq i<j \leq k^{h}$.

    \item For any objective $f \in \{f_1, f_2, \cdots, f_M \} \bigcup \{\overline{f}\} $, it satisfies that for all $x, y \in \mathcal{X}$, we have 
    \begin{equation}
    \nonumber
        f^* - f(y) \leq f^*- f(x) + \max \left \{f^*- f(x), \ell(x, y) \right \}
    \end{equation}
\end{list1}
\end{assumption}

\begin{remark}
In other words, we assume that all the local objectives as well as the global objective satisfy the smoothness property. Similar to the existing works on the $\mathcal{X}$-armed bandit problem, our proposed algorithm \texttt{PF-PNE} does not need the dissimilarity function $\ell$ as an explicit input. Only the smoothness constants $\nu_1, \rho$ are used in the objective \citep{bubeck2011X, azar2014online}. As mentioned by \citet{bubeck2011X, Grill2015Blackbox, Li2022Federated}, most regular functions satisfy Assumption \ref{assumption: local_smoothness} on the standard equal-sized partition with accessible $\nu_1$ and $\rho$. 
\end{remark}

Next, we present the additional assumption(s) on the similarity among the local objectives for the benefit of federated learning.
\begin{assumption}
\label{assumption: difference_in_optimal_values}
    \textbf{\upshape (Difference in Optimal Values)}. The global optimal value and the local optimal values are bounded by some (known) constant $\Delta$, i.e., $\forall m \in [M]$, the following property is satisfied
    \begin{equation}
        \nonumber
        |\overline{f}^* - f_m^*| \leq \Delta 
    \end{equation}
\end{assumption}
\begin{remark}
Assumption \ref{assumption: difference_in_optimal_values} is very weak since we only want an upper bound on the difference in the optimal values of  $\overline{f}^*$ and all the $f_m^*$. Note that all the objectives are bounded, therefore we always have $\Delta \leq 1$. However, setting a very large $\Delta$ would basically mean that we have no prior knowledge of the similarity between the local objectives and they can be very different. 
% Our proposed \texttt{PF-PNE} algorithm can still converge to the local optimums in this case, but it would jump to the second stage early and perform local finetuning that disables the benefit of federated learning. Although such a phenomenon is expected when the local objectives are different, our algorithm behaves much nicer when $\Delta$ is small and thus the local objectives are similar. We will first use Assumption \ref{assumption: difference_in_optimal_values} to present a general theorem, and then Assumption \ref{assumption: near_optimal_similarity} to show the benefit of \texttt{PF-PNE} when the functions are similar.
\end{remark}
\begin{assumption}
\label{assumption: near_optimal_similarity}
    \textbf{\upshape (Near-optimal Similarity)}. Let $\Delta$ be the upper bound on the difference in local and global optimum in Assumption \ref{assumption: difference_in_optimal_values}. At any $\epsilon$-near-optimal point of the global objective $\overline{f}$, the local objective is at least $(\omega \epsilon + \Delta)$-near-optimal for some $\omega \geq 1$, i.e.,  if $x \in \mathcal{X}, \epsilon > 0$ satisfies $\overline{f}^* - \overline{f}(x) \leq \epsilon$, then
    \begin{equation}
        \nonumber
        {f}_m^* - f_m(x) \leq \Delta + \omega \epsilon, \forall m \in [M]
    \end{equation}
\end{assumption}

\begin{remark}
Note that Assumption \ref{assumption: near_optimal_similarity} is also mild because we only require near-optimal points in the global objective to be near-optimal on local objectives, with the optimality difference thresholded by the number $\Delta$ {and the factor $\omega$}. Such an assumption also makes sense in real life. For example, when we tune hyper-parameters on machine learning models, if one set of hyper-parameters achieves 0.8 reward (e.g., accuracy) globally on average, then we should expect that the reward for the same set of hyper-parameters on the local objectives is not too bad, say, less than 0.7. Compared with assumptions that require everywhere similarity, e.g., $|\overline{f}(x) - f_m(x)| \leq \epsilon$ for every $x \in \mathcal{X}$ and some small $\epsilon > 0$, Assumption \ref{assumption: near_optimal_similarity} is obviously much weaker. 
% Discussions on this assumption can be found in Section \ref{sec: conclusions}
\end{remark}

%% file: Sections/03.Algorithm.tex
\section{Algorithm and Analysis}
\label{sec: algorithm}

\textbf{Challenges.} The personalized federated $\mathcal{X}$-armed bandit problem encounters several new challenges. First of all, instead of optimizing the \textit{global objective} as in \citet{shi2021federateda, Li2022Federated}, the algorithms need to optimize all the local objectives of the clients at the same time, which is obviously much harder. Besides, we have no assumptions on the difference between the local objectives except for Assumption \ref{assumption: near_optimal_similarity}. Therefore, the algorithm needs to adapt to the heterogeneity in the local objectives, i.e., more collaborations should be encouraged when they are similar, and reckless collaborations should be prevented when they are different.
Last but not least, the federated learning setting implies that only limited information, e.g., summary statistics such as the empirical average of the rewards, can be shared across the clients within limited communication rounds.

We first introduce a naive but insightful idea to solve the personalized federated $\mathcal{X}$-armed bandit problem,  which inspires our final algorithm. The approach is quite straightforward: we can ask the clients to collaboratively eliminate some regions of the domain $\mathcal{X}$, and narrow the search region of the local optimums down to a smaller subdomain. The clients can then continue the learning process by restricting their domain to the small subdomain instead of the original $\mathcal{X}$. For example, if the original parameter domain is $\mathcal{X} = [0, 1]$, the clients could first collaboratively learn the ``near-optimal" subdomain that is good for every client, say $ \mathcal{X}' = [0.5, 0.6]$, i.e., $f_m(x) - f_m^*$ is small for every $m$ on $\mathcal{X}'$. Then the clients can individually finetune the subdomain to find their local optimums. However, such an approach suffers from the problem of which region could be safely eliminated, since the local objectives could be very different on the non-near-optimal regions of $\overline{f}$ (and thus not breaking Assumption \ref{assumption: near_optimal_similarity}). On one hand, if the optimum of any client is eliminated in the collaborative learning process, then linear regret will be induced. On the other hand, if we can only eliminate a small part from $\mathcal{X}$, then the collaboration between the clients will simply be ineffective.

\subsection{The \texttt{PF-PNE} Algorithm}

\begin{algorithm}
   \caption{ \texttt{PF-PNE: server}}
   \label{alg: server}
\begin{algorithmic}[1]
   \STATE \textbf{Input:} $k$-nary partition $\mathcal{P}$, smoothness parameters $\nu_1, \rho$, transition layer $H_0$
   \STATE \textbf{Initialize} $ \mathcal{K}^1 =  \{(0, 1)\}, h = 0$
   \WHILE{not reaching the time horizon $T$}
    \STATE Update $h = h+1$
    % \WHILE{$|\mathcal{K}^{p}|\tau_{h}\leq M$}
    %         \STATE $\mathcal{K}^{p} = \left\{(h'+1, ki-j) \mid \forall (h',i) \in {\mathcal{K}}^p, j \in [k-1] \right\}$ 
    %         \STATE Renew  $h = h+1$ 
    %     \ENDWHILE
   \IF{$h \leq H_0$}
        \STATE  Receive local estimates $\{\widehat{\mu}_{m, h,i}\}_{m \in [M], (h,i) \in \mathcal{K}^p}$ from all the clients
    
        \FOR{every $(h, i) \in \mathcal{K}^h$  }
            \STATE Calculate the global mean estimate $\overline{\mu}_{h,i} = \frac{1}{M} \sum_{m=1}^M \widehat{\mu}_{m, h,i}$
        \ENDFOR
        \STATE Compute $(h,i^p) = \arg\max_{(h, i) \in \mathcal{K}^h}\overline{\mu}_{h,i} $
        \STATE Compute $\mathcal{E}^h = \left \{(h, i) \in \mathcal{K}^h \mid
        \right. $ \\ \quad $ \left. % comment if single column
        \overline{\mu}_{h,i} + b_{h,i} + \nu_1\rho^h < \overline{\mu}_{h,i^p} - b_{h,i^p}  \right \}$ 
        \STATE Update $\mathcal{K}^{h} = \mathcal{K}^{h} \backslash \mathcal{E}^{h}$
        \STATE  Broadcast the new set $\mathcal{K}^h$ and the statistics $\{\overline{\mu}_{h,i}, b_{h,i}\}_{(h,i) \in \mathcal{K}^h}$ to every client $m$. 
        \STATE Compute $ \mathcal{K}^{h+1} = \left\{(h+1,ki-j) \mid
        \right. $ \\ \quad $ \left. % comment if single column
        (h,i) \in (\mathcal{K}^h), j \in [k-1] \right\}$
    % \ELSE
    %     \STATE No Broadcast / Broadcast $\mathcal{K}^{p} = \emptyset$ to every client $m$. 
    \ENDIF
   \ENDWHILE
\end{algorithmic}
\end{algorithm}

\textbf{Algorithm Details and Double Elimination.} Based on the above preliminary idea and its potential issues, we propose the new Personalized-Federated-Phased-Node-Elimination (\texttt{PF-PNE}) algorithm that has a unique \textit{double-elimination strategy} so that collaboration among the clients are encouraged while no nodes would be readily removed without careful checks. The algorithm details are shown in Algorithms \ref{alg: client}, \ref{alg: personalize_elimination}, and \ref{alg: server}. In these algorithms, we have indexed the nodes using their depths and indices $(h,i)$. $T_{m,h,i}$ denotes the numbers of pulled samples on the $m$-th local objective in $h,i$th node and $b_{m, h, i} =  c \sqrt{\frac{\log(c_1 T/\delta)}{T_{m, h,i}}} $ is the corresponding confidence bound. $T_{h,i} = \sum_{m=1}^M T_{m, h,i}$ and $b_{h,i}$ is the confidence bound defined similarly with respect to the global objective. The details of all notations can be found in Appendix \ref{app: notaions_and_lemmas}.

Each client maintains two sets of ``active" nodes at each depth, $\mathcal{K}^{h}$ and $\mathcal{K}_m^{h}$, i.e., they are the sets of nodes that we believe potentially contains the global optimum and the local optimum respectively. $\mathcal{K}^h$ is the set of global active nodes and controlled by the server, the algorithm tries to explore the global object $\overline{f}$ over the region $\mathcal{K}^h$ collaboratively by all clients. $\mathcal{K}_m^h$ is the set of local active nodes and controlled by the client, the algorithm tries to explore the local objective $f_m$ over the region $\mathcal{K}_m^h\backslash\mathcal{K}^h$ locally by the $m$-th client.
At each depth $h$, $\mathcal{K}^h \subseteq \mathcal{K}_m^h$.
Similarly, the server and the clients also maintain two sets of nodes $\mathcal{E}^h$, $\mathcal{E}_m^h$ to be eliminated/removed from $\mathcal{K}^h$ and $\mathcal{K}_m^h$ respectively.
Each node is pulled/evaluated either globally or locally for at least $\tau_h := \left\lceil {c^2   \nu_1^{-2}\log(c_1T/\delta)}  \rho^{-2h} \right \rceil$ times for an accurate estimation of the reward, where  $c$ is an absolute constant and $\delta$ is the confidence parameter.

\begin{algorithm}
   \caption{ \texttt{PF-PNE: $m$-th client}}
   \label{alg: client}
\begin{algorithmic}[1]
\STATE \textbf{Input:} $k$-nary partition $\mathcal{P}$, smoothness parameters $\nu_1, \rho$, transition layer $H_0$
\STATE \textbf{Initialize} $h = 0$, $H_0 := \text{argmax}_{h \in \mathbb{N}} \left(\nu\rho^h \leq \Delta\right) $, $\mathcal{K}_m^0 =$the first broadcast $\mathcal{K}^h$ from the server.
\IF{$h \leq H_0$}
    \WHILE{not reaching the time horizon $T$}
        \FOR{each $(h,i) \in \mathcal{K}^h$ sequentially}
            \STATE Pull the node $\lceil\frac{\tau_h}{M} \rceil$ times and receive rewards $\{r_{m, h, i, t}\}$
        \ENDFOR
        \STATE Calculate $\widehat{\mu}_{m, h,i} = \frac{1}{ T_{m, h, i}} \sum_{t} r_{m, h, i, t}$  for every $(h,i) \in \mathcal{K}^h$
        \STATE Send the local estimates $\{\widehat{\mu}_{m, h,i}\}_{(h,i) \in \mathcal{K}^h}$ to the server
        \STATE Receive new ${\mathcal{K}}^{h}$ and $\{\overline{\mu}_{h,i}, b_{h,i}\}_{(h,i) \in \mathcal{K}^h}$
        \STATE Update $\widehat{\mu}_{m, h,i} = \overline{\mu}_{h,i}, b_{m, h,i} = b_{h,i}$ for every node s$(h,i) \in \mathcal{K}^h$
        \STATE Compute $ \mathcal{K}^{h+1} = \left\{(h+1,ki-j) \mid \right . $ \\
        \quad $\left. (h,i) \in \mathcal{K}^h, j \in [k-1] \right\}$
        \STATE Update $h = h + 1$
    \ENDWHILE
\ELSE
    \STATE Set $\mathcal{K}^h = \emptyset$ for all $h > H_0$
    \STATE Run \texttt{PE($\mathcal{P}, \nu_1, \rho$)}
\ENDIF
\end{algorithmic}
\end{algorithm}

The algorithm consists of two stages, while each stage has several phases. The depth $H_0$ for transitioning between the two stages is $H_0 := \text{argmax}_{h \in \mathbb{N}} \left(\nu\rho^h \leq \Delta\right) $, or equivalently $H_0 = - \lceil \log_{\rho}(\Delta/\nu) \rceil$. This implies that when we have control over the optimality difference $\Delta$, we can optimize the local objectives jointly, and when the control is lost, the clients should find their local optimums separately.

\begin{list1}
\item In the first stage, collaboration among the clients is encouraged to accelerate the learning process. At each depth $h$ starting from the root $\mathcal{K}^0 = {(0, 1)}$, the server sends the new $\mathcal{K}^h$ to the clients. The clients evaluate the nodes and send their average rewards $\widehat{\mu}_{m, h, i}$ back to the server. The server will then compute the global average $\overline{\mu}_{h,i}$, select the best node $\mathcal{P}_{h,i^p}$ (in terms of the $\overline{\mu}_{\cdot,\cdot}$ value),  and determine the set of nodes $\mathcal{E}^{h}$ to be eliminated by using the elimination criterion $\overline{\mu}_{h,i} + b_{h,i} + \nu_1\rho^h < \overline{\mu}_{h,i^p} - b_{h,i^p}$. The updated set $\mathcal{K}^h$ and the global average rewards of the nodes inside the set are then communicated to the clients. For nodes still inside $\mathcal{K}^h$, their local statistics $\widehat{\mu}_{m, h,i}, b_{m, h,i}$ are replaced by the global ones $\overline{\mu}_{h,i}, b_{ h,i} $

\item In the second stage, the server terminates the collaboration and the clients initiate the Personalized Elimination (\texttt{PE}) algorithm. Starting again from the root, they evaluate the nodes that are in the set $\mathcal{K}^h_m \setminus \mathcal{K}^h$ until at least $\tau_h$ local samples are obtained, including the nodes that are eliminated in the first stage ($\mathcal{E}^h$). They will then find the best node in $\mathcal{K}_m^h$ and determine the set of bad nodes $\mathcal{E}^h_m$ with a similar elimination criterion as the first stage. For all the nodes inside $\mathcal{E}^h_m$, they can now be safely removed because they have been eliminated twice, ergo double elimination, once from $\mathcal{K}^h$ and once from $\mathcal{K}_m^h$. At the same time, $\mathcal{K}^h$ will be protected from elimination and no further exploration is needed for the nodes in this set. 
\end{list1}

\begin{algorithm}
   \caption{ \texttt{PE: $m$-th client}}
   \label{alg: personalize_elimination}
\begin{algorithmic}[1]
\STATE \textbf{Input:} partition $\mathcal{P}$, smoothness parameters $\nu_1, \rho$
\STATE \textbf{Initialize} $h = 0$, $\mathcal{K}_m^0 = \{(0, 1)\}$.
\WHILE{not reaching the time horizon $T$}
        \WHILE{$T_{m,h,i} < \tau_h $ for any  $(h, i) \in \mathcal{K}_m^h \backslash  \mathcal{K}^h$ }
            \STATE Pull the node and receive reward $r_{m, h, i, t}$ 
        \ENDWHILE
        % \WHILE{$|\mathcal{T}_2^p| + |\mathcal{T}_3^p|$ rounds are not reached}
        %     \STATE Randomly pull a node in $\mathcal{K}_m^p$ and receive reward  $r_{m, h, i, t}$ 
        % \ENDWHILE
        \STATE Calculate $\widehat{\mu}_{m, h,i} = \frac{1}{ T_{m, h, i}} \sum_{t} r_{m, h, i, t}$  for every $(h,i) \in \mathcal{K}_m^h \backslash  \mathcal{K}^h$
        \STATE Compute $(h, i_m^p) = \arg\max_{(h, i) \in \mathcal{K}_m^h}\widehat{\mu}_{m, h,i} $
        \STATE Compute  $\mathcal{E}_m^h = \left \{(h, i) \in \mathcal{K}_m^h \backslash  \mathcal{K}^h \mid
        \right. $ \\ \quad $ \left. % comment if single column
        \widehat{\mu}_{m, h,i} + b_{m, h,i} + \nu_1\rho^h < \widehat{\mu}_{h,i_m^p} - b_{h, i_m^p}  \right \}$  
        \STATE Compute $ \mathcal{K}_m^{h+1} = \left\{(h+1,ki-j) \mid 
        \right. $ \\ \quad $ \left. % comment if single column
        (h,i) \in (\mathcal{K}_m^h \setminus \mathcal{E}_m^h), j \in [k-1] \right\}$
        \STATE Update $h = h+1$
\ENDWHILE
\end{algorithmic}
\end{algorithm}

\begin{remark}
    In Algorithm \ref{alg: client} and \ref{alg: personalize_elimination} for client $m$, ``pulling a node $\mathcal{P}_{h,i}$" refers to evaluating the local objective $f_m$ at a particular point $x \in \mathcal{P}_{h,i}$ in order to obtain the reward. Note that we assume local smoothness on all the objectives (Assumption \ref{assumption: local_smoothness}), therefore whether we randomly choose the evaluation point inside the node for each evaluation, or use one pre-determined point for all nodes does not affect the final regret bound. For simplicity, we choose the latter design in our analysis and our experiments. Similar results are observed in \citep{bubeck2011X, azar2014online, Li2022Federated}.
\end{remark}

\textbf{Algorithm Uniqueness.} The uniqueness of the design in \texttt{PF-PNE} is four-fold: 
\begin{list1}
    \item \textbf{(Collaboration)}. At each depth $h$, the number of samples needed for the nodes in $\mathcal{K}^h$ is reduced by the collaboration between the clients, and thus making the per-client cumulative regret smaller than single-client algorithms on these nodes.
    
    \item \textbf{(Double Elimination)}.  The double-elimination strategy guarantees that the nodes are safely eliminated so that the optimum of any local objective will not be directly removed, and thus protecting the cumulative regret from being linear (See Theorem \ref{thm: regret_upper_bound} and Remark \ref{rmk: main_theorem_remark}). 
    
    \item \textbf{(Biased Evaluation Helps)}. In the second stage, the clients will utilize the global average reward and confidence bound information on $\mathcal{K}^h$ in the first stage to perform the second elimination process. No further exploration or eliminations will be performed on those nodes. On one hand, this strategy essentially reduces the sampling cost for the nodes in $\mathcal{K}^h$ for every client. On the other hand, despite that the global average $\overline{\mu}_{h,i}$ of the rewards is a biased evaluation of the local objective $f_m$ at node $\mathcal{P}_{h,i}$, we could still use it to substitute the local average $\widehat{\mu}_{m, h,i}$. As we show in the analysis, the size of the bias is under control and the biased evaluations are still helpful. 
    
    \item \textbf{(Limited Communications)} Based on our design and the choice of the stage transitioning criterion, the communication cost is always limited, both in terms of rounds and information, which makes sure that no frequent communications between the server and the clients are needed. 
\end{list1}

\subsection{Theoretical Analysis}

In order to analyze the cumulative regret of the proposed algorithm, we introduce the definition of the near-optimality dimension, which is a common notation in the existing literature that measures the number of near-optimal regions and thus the difficulty of the problem \citep{bubeck2011X, azar2014online, shang2019general, li2022communication}.

\begin{definition}
\label{definition: near-optimality dimension}
{\upshape \textbf{(Near-optimality Dimension)}}
Let $\epsilon_h >0 $ and $\epsilon'_h > 0$ be two functions of $h$, for any subset of $\epsilon_h$-optimal nodes for the function $f$, $\mathcal{X}_{f, \epsilon_h} = \{ x\in \mathcal{X} : f^* - f(x) \leq \epsilon_h \}$, there exists a constant $C$ such that $\mathcal{N}_f(\epsilon_h, \epsilon'_h) \leq C (\epsilon'_h)^{-d}, \forall h \geq 0$, where $d:= d_f(\epsilon_h, \epsilon'_h)$ is the near-optimality dimension of the function $f$ and $\mathcal{N}_f(\epsilon_h, \epsilon_h')$ is the $\epsilon_h'$-cover number of the set $\mathcal{X}_{f, \epsilon_h}$ w.r.t. the dissimilarity $\ell$.
\end{definition}

% \begin{remark}
% \end{remark}

\label{subsec: theoretical_analysis}
Using the above near-optimality dimension definition, we denote $d_m = d_{f_m}(12\nu\rho^h, \rho^h)$ for every $m \in [M]$ and { $\overline{d} = {d}_{\overline{f}}(6\nu\rho^h, \rho^h)$ }. Define $d_{\max} = \max \{d_1, d_2, \cdots, d_M\}$ and $d_{\min} = \min \{d_1, d_2, \cdots, d_M\}$.  Now we provide the general upper bound using near-optimality dimension, on the cumulative regret of the proposed \texttt{PF-PNE} algorithm as follows.

\begin{theorem}
\label{thm: regret_upper_bound}
Suppose that all the local objectives $f_1, f_2 \cdots, f_M$ and the global objective $\overline{f}$ all satisfy Assumptions \ref{assumption: local_smoothness}, \ref{assumption: difference_in_optimal_values}. Setting $\delta = 1/M$ in Algorithm \ref{alg: client}, \ref{alg: personalize_elimination}, and \ref{alg: server}, the expected cumulative regret of the \texttt{PF-PNE} algorithm satisfies
\begin{equation}
\begin{aligned}
\nonumber
  \mathbb{E}[R (T)] 
& = \widetilde{\mathcal{O}} \left( M^\frac{\overline{d} + 1}{\overline{d} + 2} T^\frac{\overline{d}  + 1}{\overline{d} + 2} + M T^\frac{{d}_{\max}  + 1}{{d}_{\max}  + 2}\right)\\
\end{aligned}
\end{equation}
The number of communication rounds of \texttt{PF-PNE} scales as $\min\{C_1 M \log \frac{1}{\Delta},  C_2(M \log MT) \}$, where $C_1$, $C_2$ are two absolute constants.
\end{theorem}

\begin{remark}
\label{rmk: main_theorem_remark}
We relegate the proof of the above theorem to Appendix \ref{app: Main_Proof}. We emphasize that since we only use Assumption \ref{assumption: difference_in_optimal_values} without any requirements on the size of $\Delta$, the regret upper bound in Theorem \ref{thm: regret_upper_bound} displays a preliminary and natural result. 

The regret bound consists of two terms
\begin{list1}
    \item The first term $\widetilde{\mathcal{O}} \left( M^\frac{\overline{d} + 1}{\overline{d} + 2} T^\frac{\overline{d}  + 1}{\overline{d} + 2} \right)$ comes from the first stage of elimination in \texttt{PF-PNE}, which might continue for a  large number of rounds if $\Delta$ is very small. In that case, the federated learning process could be viewed as $M$ clients optimizing $\overline{f}$ jointly, and thus the regret is related to the near-optimality dimension $\overline{d}$  of $\overline{f}$.  
    \item The second term $\widetilde{\mathcal{O}} \left(M T^\frac{{d}_{\max}  + 1}{{d}_{\max}  + 2}\right)$ comes from the second round of elimination in \texttt{PF-PNE}.  When $\Delta$ is large, e.g., $\Delta = 1$, it implies that we have almost no prior beliefs on the difference between the local and global optimum values. In that case, the local objectives could be very different and we can only bound the cumulative regret asymptotically by the objective with the largest near-optimality dimension, or equivalently, the hardest local objective. 
\end{list1}

Note that both two terms are sublinear with respect to $T$, it means that \texttt{PF-PNE} is always capable of finding the optimums of the local objective, regardless of whether the prior knowledge of $\Delta$ is small or large. Therefore, we claim that \texttt{PF-PNE} ``always works".

In order to analyze the regret more tightly, we introduce the following new notation $d_{\text{new}}$ to measure the difference in local and global optimal nodes, and thus the size of $\mathcal{K}^h_m \setminus \mathcal{K}^h$. This set $\mathcal{K}^h_m \setminus \mathcal{K}^h$ is, in the worst case, $\Omega (\rho^{-d_{m}h})$ at each depth $h$, such as when we terminate the first stage early with large $\Delta$ and $\mathcal{K}^h$ is simply empty, but it should be much smaller when the objectives are similar. Moreover, we need to assume a reasonably small $\Delta$ for \texttt{PF-PNE} to outperform single-client algorithms.
\end{remark}

\begin{definition}
\label{definition: optimality-difference dimension}
{\upshape \textbf{(Optimality-Difference Dimension)}}
Using the same notations as in Definition \ref{definition: near-optimality dimension}, the optimality-difference dimension  is defined to be the smallest number $d_{\text{new}} \geq 0$ such that $\mathcal{N}_{f_m}(12\nu\rho^h, \nu\rho^h) \setminus \mathcal{N}_{\overline{f}}(6\nu\rho^h, \rho^h) \leq C_0 \rho^{-d_{\text{new}}h}, \forall m \in [M]$.
\end{definition}

\begin{remark}
First of all, the above definition could be defined with respect to each client, but asymptotically we would care about the largest one across all the clients. 
Based on the definition of optimality-difference dimension, we know that $d_{\text{new}} \leq d_{\max}$ and it is a tighter measure of the number of local near-optimal nodes that are non-optimal globally.
Using the definition, we provide the following corollary as a tighter upper bound on the cumulative regret.
\end{remark}

\begin{corollary}
\label{corollary: regret_upper_bound}
Suppose that all the assumptions in Theorem \ref{thm: regret_upper_bound} and Assumption \ref{assumption: near_optimal_similarity} are satisfied, and  $d_{\text{new}}$ is the optimality-difference dimension as in Definition \ref{definition: optimality-difference dimension}. Assume that $\Delta \leq C_3 (\log MT / T)^{\frac{1}{\max\{d_{\min}, d_{\text{new}}\}+2}}$, where $C_3$ is an absolute constant, the expected cumulative regret of the \texttt{PF-PNE} algorithm satisfies
\begin{equation}
\begin{aligned}
\nonumber
  \mathbb{E}[R (T)] 
& =\widetilde{\mathcal{O}} \left( M^\frac{d_{\min}  + 1}{d_{\min}   + 2} T^\frac{d_{\min}   + 1}{d_{\min} + 2} + M T^\frac{d_{\text{new}}  + 1}{d_{\text{new}}  + 2}\right)\\
\end{aligned}
\end{equation}
\end{corollary}

\begin{remark}
 When the local objectives are similar and most local near-optimal nodes are also globally near-optimal, then $d_{\text{new}} \ll {d}_{\min}$ and the above regret bound will be dominated by the first term. It means that on average, the client-wise regret is of order $ \widetilde{\mathcal{O}} \left( M^{-\frac{1}{d_{\min}  + 2} }T^\frac{d_{\min}  + 1}{d_{\min}  + 2}\right)$, then the cumulative regret will be smaller than running the $\mathcal{X}$-armed bandit algorithms (e.g., \citet{azar2014online, li2021optimumstatistical}) separately on the clients. Similar to the arguments in Remark \ref{rmk: main_theorem_remark}, when the local objectives are different, $d_{\text{new}}$ will be almost the same as $d_{\max}$ and running the \texttt{PF-PNE} algorithm will be asymptotically the same as running the $\mathcal{X}$-armed bandit algorithms. Therefore, the proposed \texttt{PF-PNE} algorithm adapts to the heterogeneity of the local objectives.
\end{remark}

\begin{figure*}
\centering
\hspace*{-1em}
\subfigure[\footnotesize Garland (1D)]{
  \centering
  \includegraphics[width=0.27\linewidth]{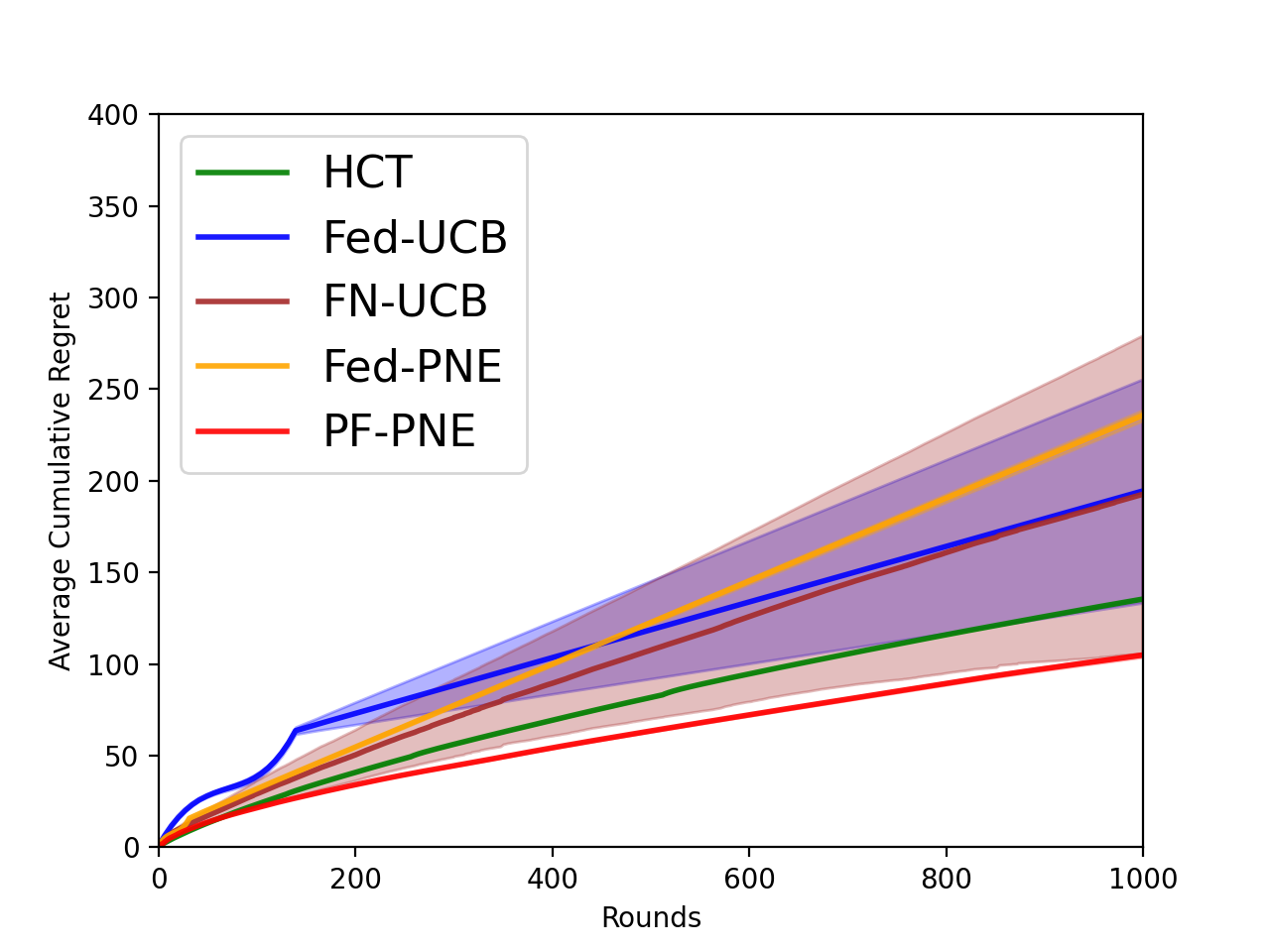}
  \label{fig: Garland}
}\hspace*{-1.5em}%
\subfigure[\footnotesize Himmelblau (2D)]{
  \centering
  \includegraphics[width=0.27\linewidth]{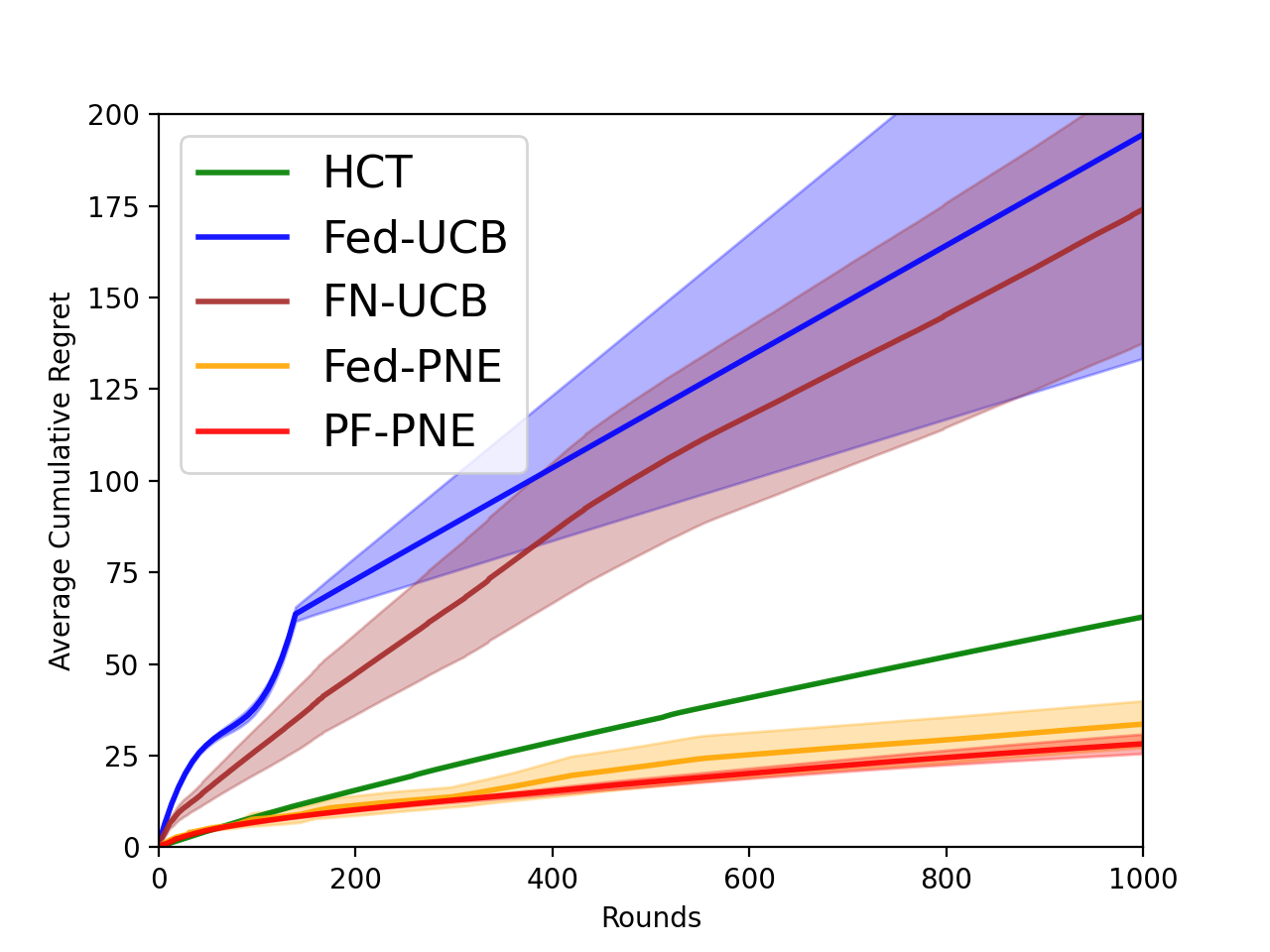}
  \label{fig: Himmelblau}
}\hspace*{-1.5em}%
\subfigure[\footnotesize Rastrigin (10D)]{
  \centering
  \includegraphics[width=0.27\linewidth]{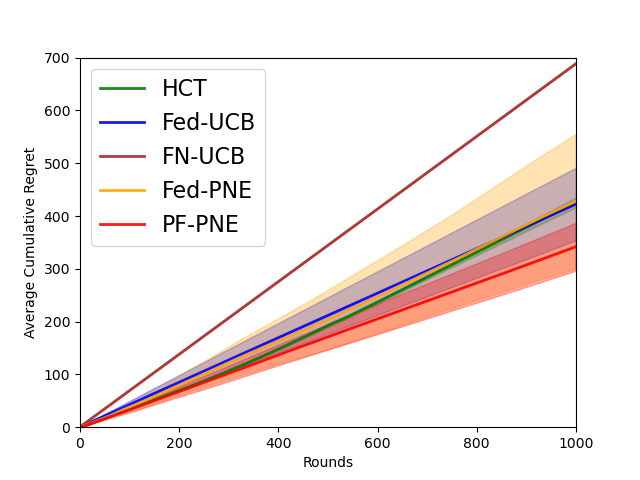}
  \label{fig: Rastrigin}
}\hspace*{-1.5em}%
\subfigure[\footnotesize Landmine]{
  \centering
  \includegraphics[width=0.27\linewidth]{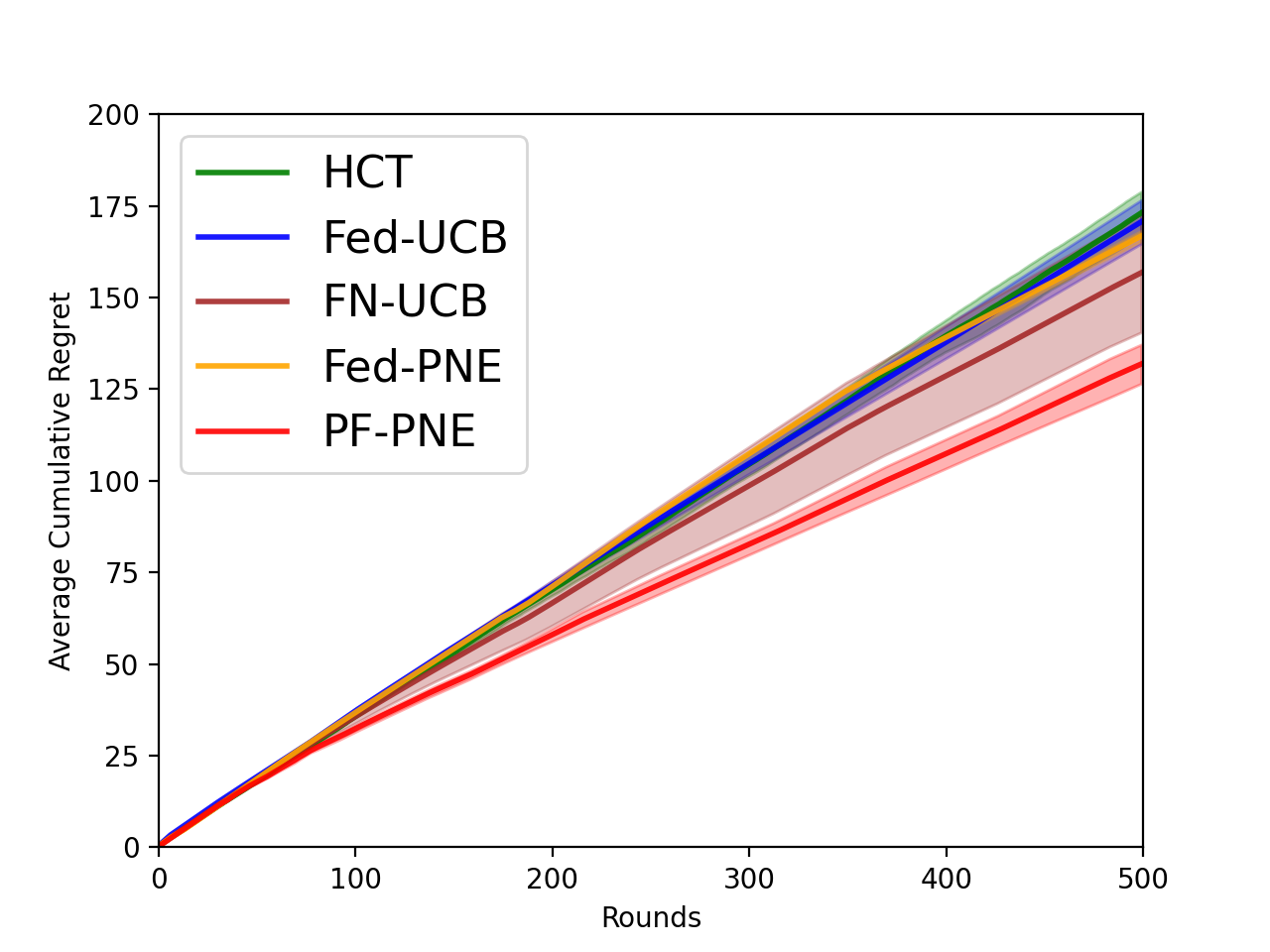}
  \label{fig: Landmine}
}
\vspace{-10pt}
\caption{Cumulative regret of different algorithms on the synthetic functions and the real-life datasets. Unlimited communications are allowed for centralized algorithms. \textbf{\texttt{FNUCB} runs much slower compared with the other algorithms due to the training of neural nets. The other four algorithms take similar time.}}
\vspace{-10pt}
\label{fig: experiments}
\end{figure*}

\begin{remark}
\textbf{(Communication Cost)} 
    The number of communication rounds in \texttt{PF-PNE} is always bounded by the minimum between a constant that depends on $\Delta$ in Assumption \ref{assumption: difference_in_optimal_values} and a term that depends on $T$ logarithmically. 
    \begin{list1}
    \item When $\Delta > 0$ is slightly large (e.g., 0.1) and $T$ is sufficiently large, the communication cost is always bounded by a constant $C_1 M \log \frac{1}{\Delta}$. The bound makes sense intuitively because as long as the local objectives are different, we should ask the clients to find their local optimums by themselves and further communications become futile at some point in the learning process. Therefore, both the number of communication \textbf{rounds} and the amount of \textbf{information} communicated will be bounded.

    \item  If $\Delta$ is a very small number, or even the extreme case 0, and the term $ C_2(M \log MT)$ dominates the communication cost, it means that the \texttt{PF-PNE} algorithm will degenerate to almost the same as \texttt{Fed-PNE} in \citet{Li2022Federated}, because the second stage will not be activated and the second round of elimination will never be executed. In this case, the communication cost would be the same as \texttt{Fed-PNE}. \citet{Li2022Federated} have proved that the amount of \textbf{information} transferred would be of order $\widetilde{\mathcal{O}}(M \log T\vee MT^{\frac{d}{d+2}})$, which is still sublinear. Such a communication cost is proven to be unavoidable \citep{Li2022Federated}.
    \end{list1}
    
     The extreme case $\Delta = 0$ could happen only if the local objectives have the same optimum at the same point. In all our experiments, we observe that the communication cost is always bounded.
\end{remark}

%% file: Sections/04.Experiments.tex
\section{Experiments}
\label{sec: experiments}

In this section, we provide the empirical evaluations of the proposed \texttt{PF-PNE} algorithm on both synthetic and real-world objectives. We compare \texttt{PF-PNE} with \texttt{HCT} \citep{azar2014online}, \texttt{Fed1-UCB} \citep{shi2021federateda}, \texttt{FN-UCB} \citep{dai2023federated} and \texttt{Fed-PNE} \citep{Li2022Federated}. The curves in the figures are averaged over 10 independent runs of each algorithm with the shaded regions representing 1 standard deviation error bar. Additional experimental details and algorithm implementations can be found in Appendix \ref{app: experimental_details}.
\begin{remark}
    For \texttt{HCT}, we run the algorithm on the $M$ local objectives and then plot the average local regret across all the objectives with no communications. For the other federated algorithms, we also plot the average cumulative regret across the clients. Such a comparison is fair since we essentially compare the single-client algorithms with the federated algorithms on their average performance across multiple clients.
\end{remark}

\textbf{Synthetic Objectives. }
We first conduct experiments on three synthetic objectives, Garland, Himmelblau, and Rastrigin, with the parameter domains $\mathcal{X}$ to be $[0, 1]$,  $[-5, 5]^2$, and $[-1, 1]^{10}$ respectively. We apply random shifts to the original synthetic functions to create the local objectives. The shifts are zero-mean normal random variables and are applied to every dimension of the objective. The average cumulative regrets of different algorithms are provided in Figure \ref{fig: Garland}, \ref{fig: Himmelblau}, and \ref{fig: Rastrigin}. As can be observed in the figures, 
\texttt{PF-PNE} has the smallest averaged cumulative regret. The performance of \texttt{Fed-PNE} largely depends on the similarity between the local objectives and the global objective because it is designed to optimize the global objective. When they are very different, e.g., on Garland and Rastrigin, the performance of \texttt{Fed-PNE}, although better than its competitors, is far from satisfactory.

\textbf{Landmine Detection}. The landmine dataset \citep{liu2007semi} consists of multiple landmine fields with different locations of the landmine extracted from radar images. For each client, we randomly assign one of the landmine fields to the client. We federatedly tune the hyper-parameters of support vector machines with the RBF kernel parameter chosen from $[0.01, 10]$ and the $L_2$ regularization parameter chosen from $[10^{-4}, 10]$. The local objectives are the AUC-ROC scores of the support vector machine evaluated on the local landmine fields. We provide the average cumulative regret of different algorithms in Figure \ref{fig: Landmine}. As shown in the figure, our algorithm achieves the smallest regret.

%% file: Sections/05.Conclusions.tex
\section{Discussions and Conclusions}
\label{sec: conclusions}

In this work, we study the personalized federated $\mathcal{X}$-armed bandit problem and propose the first algorithm for such problems. The proposed \texttt{PF-PNE} algorithm utilizes the hierarchical partition and the idea of double elimination to help the clients locate their own optimums. \texttt{PF-PNE} is unique in its adaptivity to the heterogeneity of the local objectives and its little communication cost for the federated learning process. Several interesting future directions are also inspired by our work. For example, is our similarity measure $\Delta$ the best assumption to quantify the difference between the local objectives, or is there an even weaker/ more useful assumption for personalized federated $\mathcal{X}$-armed bandit? 
% For example, in our analysis, Assumption \ref{assumption: near_optimal_similarity} can be even weakened to the following assumption. 
% %
% %
% With only a few changes to our proofs, we are able to provide a regret bound for the PF-PNE algorithm $ \widetilde{\mathcal{O}} \left( M^\frac{\overline{d} + 1}{\overline{d}  + 2} T^\frac{\overline{d} + 1}{\overline{d}  + 2} + M T^\frac{{d}_{\max}  + 1}{{d}_{\max}  + 2}\right)$ similar to Theorem \ref{thm: regret_upper_bound}, where $\overline{d}$ denotes the near-optimality dimension of $\overline{f}$.
% However, with this even weaker assumption, we are unable to show the relationship between $\overline{d}$ and $d_{\max}$, or equivalently the relationship between the number of near-optimal nodes in $\overline{f}$ and $\{f_1, f_2, \cdots, f_M\}$.  
Besides, \texttt{PF-PNE} still needs the smoothness parameters and the prior knowledge on the bound $\Delta$ as part of the input, and it would be interesting to explore parameter-free algorithms in our setting.

%% file: Sections/Appendix01.tex
\section{Notations and Useful Lemmas}
\label{app: notaions_and_lemmas}

\subsection{Notations}

Here we list all the notations used in the proof of our cumulative regret bound:
\begin{list2}
    \item $\mathcal{L}_{t}$ denotes all the nodes in the exploration tree at time $t$
    \item $\mathcal{T}_m^h$: The time steps of client $m$ spent on layer $h$.
    \item $\mathcal{K}^h$: The set of pre-eliminated nodes in the server on layer $h$.
    \item $\mathcal{E}^h$: The set of nodes to be eliminated in the server on layer $h$.
    \item $\mathcal{K}_m^h$: The set of pre-eliminated nodes in the client $m$ on layer $h$.
    \item $\mathcal{E}_m^h$: The set of nodes to be eliminated in the client $m$ on layer $h$.
    \item $\overline{\mathcal{K}}^h$: The set of post-eliminated nodes from the server at depth $h$, i.e., $\overline{\mathcal{K}}^h = \mathcal{K}^h \setminus \mathcal{E}^h$.
    \item $\overline{\mathcal{K}}_m^h$: The set of post-eliminated nodes from the client $m$ at depth $h$, i.e., $\overline{\mathcal{K}}_m^h = \left(\mathcal{K}_m^h \setminus \overline{\mathcal{K}}^h \right) \setminus \mathcal{E}_m^h$.
    \item $(h, i^p)$: the depth and the index of the node $\mathcal{P}_{h, i^p}$ chosen by the server on layer $h$ from $\mathcal{K}^h$.
    \item $(h, i^*)$:  the depth and the index of the node $\mathcal{P}_{h, i^*}$ that contains (one of) the maximizer  $\overline{x}^*$ of the global objective $\overline{f}$ on depth $h$.
    \item $(h, i_m^{*})$:  the depth and the index of the node $\mathcal{P}_{h, i_m^*}$ that contains (one of) the maximizer $x_m^*$ of the local objective $f_m$ on depth $h$.
    \item $T_{h,i}$: the number of times the node $\mathcal{P}_{h,i}$ is sampled globally, i.e., $T_{h,i} = \sum_{m=1}^M T_{m,h,i}$.
    \item $T_{m,h,i}$: the number of times the node $\mathcal{P}_{h,i}$ is sampled from client $m$.

    \item $b_{h,i}  = c \sqrt{\frac{\log(c_1 T/\delta)}{T_{h,i}}} $: confidence bound for the node $(h,i)$ on the global objective
    \item $b_{m, h,i}  = c \sqrt{\frac{\log(c_1 T/\delta)}{T_{m, h,i}}} $: confidence bound for the node $(h,i)$ on the $m$-th lobal objective
    \item $H_t$: the maximum depth reached by the algorithms at time $t$
    \item $\tau_h$: the minimum required number of samples needed for a node on depth $h$, defined below.
\end{list2}

\textbf{The threshold for every depth.} 
The number of times $\tau_{h}$ needed for the statistical error (the UCB term) of every node on depth $h$ to be better than the optimization error is the solution to
  
\begin{equation}
    \nu_1 \rho^h \approx c \sqrt{\frac{\log(c_1 T/\delta)}{\tau_h}},
\end{equation}
  
which is equivalent as the following choice of the threshold 
\begin{equation}
\label{eqn: tau_upper_lower_bound}
  \frac{c^2 }{  \nu_1^2}  \rho^{-2h}  \leq \tau_h = \left\lceil \frac{c^2 \log(c_1T/\delta)}{  \nu_1^2}  \rho^{-2h} \right \rceil \leq 2  \frac{c^2 \log(c_1T/\delta)}{  \nu_1^2}  \rho^{-2h}.
\end{equation}
  
Notably, this choice of the threshold is the same as the threshold value in the \texttt{HCT} algorithm \citep{azar2014online}. In other words, we design our algorithm so that the samples are from different clients uniformly and thus the estimators are unbiased, and at the same time we minimize the unspent budget due to such distribution. There is still some (manageable) unspent budget due to the floor operation in the computation of $t_{m,h,i}$. However because of the expansion criterion (line 5-6) in \texttt{Fed-PNE}, we are able to travel to very deep layers inside the partition very fast when there are a lot of clients, and thus \texttt{Fed-PNE} is faster than single-client $\mathcal{X}$-armed bandit algorithms.

\subsection{Supporting Lemmas}

\begin{lemma} 
\label{lem: hoeffding}
\textbf{\upshape (Hoeffding's Inequality)}
    Let $X_{1}, \ldots, X_{n}$ be independent random variables such that $a_{i} \leq X_{i} \leq b_{i}$ almost surely. Consider the sum of these random variables, $S_{n}=X_{1}+\cdots+X_{n}$.
Then for all $t>0$, we have
\begin{equation}
\begin{aligned}
\nonumber
 \mathbb{P}\left(\left|S_{n}-\mathbb{E}\left[S_{n}\right]\right| \geq t\right) \leq 2 \exp \left(-\frac{2 t^{2}}{\sum_{i=1}^{n}\left(b_{i}-a_{i}\right)^{2}}\right).
\end{aligned}
\end{equation}
Here $\mathbb{E}\left[S_{n}\right]$ is the expected value of $S_{n}$.
\end{lemma}

\begin{lemma}
    \label{lem: good_event}
    \textbf{\upshape (High Probability Event)} At each time $t$, define the ``good" events $E_t^{1}, E_t^{2}$ as
\begin{equation}
\begin{aligned}
    &E_t^{1} = \left\{ \forall h' \leq H_t, \forall (h,i) \in \mathcal{K}^{h'}, \forall T_{h,i} \in [MT], | \overline{f}(x_{h,i}) - \overline{\mu}_{h,i}| \leq  c \sqrt{\frac{\log(c_1 T/\delta)}{T_{h,i}}} \right\} \\
    &E_t^{2} = \left\{ \forall h' \leq H_t, \forall m \in [M], \forall (h,i) \in \mathcal{K}_m^{h'} \setminus \mathcal{K}^{h'}, \forall T_{m, h,i} \in [MT], | f_m(x_{h,i}) - \widehat{\mu}_{m, h,i}| \leq  c \sqrt{\frac{\log(c_1 T/\delta)}{T_{m, h,i}}} \right\}
\end{aligned}
\end{equation}
 where the right hand sides in thee  two events are the confidence bound $b_{h,i}$ and $b_{m,h,i}$ respectively for the node $\mathcal{P}_{h,i}$ and $c \geq 2, c_1 \geq (2M^2)^{1/8}$ are two constants. Define the event $E_t = E_t^{1} \bigcap E_t^{2}$, then for any fixed round $t$, we have
$
     \mathbb{P}(E_t) \geq 1 - 2\delta/T^6
$
\end{lemma}
\textbf{Proof}.
For the first event $E_t^1$,  by utilizing the results of Lemma B.2 in \citet{Li2022Federated} and the Hoeffding's Inequality, we know that  
\begin{equation}
  \mathbb{P}\left( \left| \sum_{m \in [M]} \sum_{t \in [t_{m,h,i}]} r_{m, h, i, t} -  \sum_{m \in [M]} t_{m,h,i} f_m (x_{h,i}) \right| \geq x \right) \leq 2 \exp \left(-\frac{2 x^{2}}{T_{h,i}}\right).
\end{equation}
Therefore by the union bound, the probability of the compliment event ${E}_t^{1c}$ can be bounded as
\begin{equation}
\begin{aligned}
\mathbb{P} \left ({E}_{t}^{1\mathrm{c}}\right)
&\leq \sum_{h' \in H_t } \sum_{(h, i) \in \mathcal{K}^{h'}} \sum_{T_{h,i}=1}^{MT} \mathbb{P}\bigg(| \overline{f}(x_{h,i}) - \overline{\mu}_{h,i}| >  b_{h,i} \bigg)
\leq \sum_{h' \in H_t } \sum_{(h, i) \in \mathcal{K}^{h'}} 2MT \exp \bigg(- 2 T_{h,i} b_{h,i}^{2} \bigg)\\
& = 2MT \exp \bigg(-2c^2 \log(c_1 T/\delta) \bigg) \left ( \sum_{h' \in H_t }  |\mathcal{K}^{h'}| \right)  \leq 2 MT^2 \left(\frac{\delta}{c_1 T} \right)^{2c^2} \leq \frac{\delta}{T^6}.
\end{aligned}
\end{equation}
For the second event $E_t^{2}$, similarly we have the probability of the compliment event bounded as
\begin{equation}
\begin{aligned}
\mathbb{P} \left ({E}_{t}^{2 \mathrm{c}}\right)
&\leq \sum_{m=1}^M \sum_{h' \in H_t } \sum_{(h, i) \in \mathcal{K}_m ^{h'}} \sum_{T_{m, h,i}=1}^{MT} \mathbb{P}\bigg(| {f}_m (x_{h,i}) - \widehat{\mu}_{m, h,i}| >  b_{m, h,i} \bigg) \\
&\leq \sum_{m=1}^M \sum_{h' \in H_t } \sum_{(h, i) \in \mathcal{K}_m^{h'}} 2MT \exp \bigg(- 2 T_{m, h,i} b_{m,sh,i}^{2} \bigg)\\
& = 2M^2 T \exp \bigg(-2c^2 \log(c_1 T/\delta) \bigg) \left ( \sum_{h' \in H_t }  |\mathcal{K}_m^{h'}| \right)   \leq 2 M^2 T^2 \left(\frac{\delta}{c_1 T} \right)^{2c^2} \leq \frac{\delta}{T^6}.
\end{aligned}
\end{equation}
Finally by the union bound, we know that 
\begin{equation}
\begin{aligned}
\mathbb{P} \left ({E}_{t} \right) = 1 - \mathbb{P} \left ({E}_{t}^c \right) = 1 - \mathbb{P} \left (E_t^{1c} \bigcup E_t^{2c}\right) \geq 1 - \mathbb{P} \left (E_t^{1c} \right) -  \mathbb{P} \left (E_t^{2c} \right) \geq 1 - 2\delta/T^6
\end{aligned}
\end{equation}

\hfill $\square$

\begin{lemma}
    \label{lem: global_remaining_nodes}
    \textbf{\upshape (Optimality in Global Objective, Lemma A.4 in \citet{Li2022Federated}).} For any client $m$, under the high probability event ${E}_t$ at time $t \in \mathcal{T}_m^{h+1}$, the representative point $x_{h,i}$ of every un-eliminated node $\mathcal{P}_{h, i} $ at the previous depth, i.e., $(h,i) \in \overline{\mathcal{K}}^{h}$, is at least $6\nu_1 \rho^{h}$-optimal, that is
    \begin{equation}
        \overline{f}^* - \overline{f}(x_{h, i}) \leq 6\nu_1 \rho^{h}, \forall (h,i) \in \overline{\mathcal{K}}^{h}.
    \end{equation}
\end{lemma}
\textbf{Proof}. The proof is provided for completeness. Under the high probability event ${E}_t$, we have the following inequality for every node $\mathcal{P}_{h, i}$ such that $ (h,i) \in \overline{\mathcal{K}}^{h}$
\begin{equation}
\begin{aligned}
      | \overline{f}(x_{h, i}) - \overline{\mu}_{h, i}| \leq b_{h, i} = c \sqrt{\frac{\log(c_1 T/\delta)}{T_{h, i}}}.
\end{aligned}
\end{equation}
Therefore the following set of inequalities hold 
\begin{equation}
\begin{aligned}
       \overline{f}(x_{h, i}) + \nu_1 \rho^{h} + 2 b_{h, i} &\geq  \overline{\mu}_{h, i} + \nu_1 \rho^{h} + b_{h, i}  \geq  \overline{\mu}_{h, i^{p}} - b_{h, i^{p}} \geq  \overline{\mu}_{h, i^{*}} - b_{h, i^{p}}  \\
       &\geq  \overline{f}(x_{h, i^{*}})  - b_{h, i^{*}}  - b_{h, i^{p}} \geq  \overline{f}^* - \nu_1 \rho^{h}  - b_{h, i^{*}}  - b_{h, i^{p}},
\end{aligned}
\end{equation}
where the second inequality holds because $\mathcal{P}_{h,i}$ is not eliminated. The third inequality holds because of the elimination criterion in Algorithm \ref{alg: server}, and the last one follows from the weak lipchitzness assumption (Assumption \ref{assumption: local_smoothness}). In conclusion, we have the following upper bound on the regret
\begin{equation}
\begin{aligned}
        \overline{f}^* - \overline{f}(x_{h, i}) \leq  2\nu_1 \rho^{h}  + 2 b_{h, i}  +  b_{h, i^{*}}  + b_{h, i^{p}} \leq 6\nu_1 \rho^{h}
\end{aligned}
\end{equation}
where the last inequality holds because we sample each node enough number of times ($T_{h,i}$ larger than the threshold $\tau_h$) so that $b_{h, i} \leq \nu_1 \rho^{h}$ and thus $b_{h, i}, b_{h, i^{*}}, b_{h, i^{p}}$ are all smaller than $\nu_1 \rho^{h}$.   \hfill $\square$

\begin{lemma}
    \label{lem: local_remaining_nodes}
    \textbf{\upshape (Optimality in Local Objective)} For any client $m$, under the high probability event ${E}_t$ at time $t \in \mathcal{T}_m^{h+1}$, the representative point $x_{h,i}$ of every un-eliminated node $\mathcal{P}_{h, i}$ at the previous depth $h$, i.e., $(h,i) \in \overline{\mathcal{K}}_m^{h} \backslash \overline{\mathcal{K}}^{h}$, is at least $(11\nu_1 \rho^{h}+ \Delta)$-optimal, that is
    \begin{equation}
        {f}_m^* - {f}_m(x_{h, i}) \leq 11 \nu_1 \rho^{h} + \Delta, \forall (h,i) \in \overline{\mathcal{K}}_m^{h} \backslash \overline{\mathcal{K}}^{h} 
    \end{equation}
\end{lemma}

\textbf{Proof. } If $(h, i_m^*) \in \overline{\mathcal{K}}_m^{h} \backslash \overline{\mathcal{K}}^{h}$, i.e., the node that contains the local optimum at depth $h$, is in the set $\overline{\mathcal{K}}_m^{h} \backslash \overline{\mathcal{K}}^{h}$, then we have the following inequalities
\begin{equation}
\begin{aligned}
           {f}_m(x_{h, i}) + \nu_1 \rho^{h} + 2 b_{m, h, i} 
       &\geq  \widehat{\mu}_{m, h, i} + \nu_1 \rho^{h} + b_{m, h, i} 
       \geq  \widehat{\mu}_{m, h, i_m^{p}} - b_{m, h, i_m^{p}}  \geq  \widehat{\mu}_{m, h, i_m^{*}} - b_{m, h, i_m^{p}} \\
       &\geq  f_m(x_{h, i_m^{*}})  - b_{m, h, i_m^{*}}  - b_{m, h, i_m^{p}}  \geq  f_m^* - \nu_1 \rho^{h}  - b_{m, h, i_m^{*}}  - b_{m, h, i_m^{p}},
\end{aligned}
\end{equation}
Therefore we know that the following bound holds
\begin{equation}
\begin{aligned}
      {f}_m^* - {f}_m(x_{h, i})
      \leq 2 \nu_1 \rho^h + b_{m, h, i_m^{*}}  + b_{m, h, i_m^{p}} +  2 b_{m, h, i} 
      \leq 6 \nu_1 \rho^h
\end{aligned}
\end{equation}
where the last inequality is because $ b_{m, h, i_m^{*}},  b_{m, h, i_m^{p}}, b_{m, h, i}$ are all smaller than $\nu_1\rho^h$. On the other hand, if $(h, i_m^*) \in \overline{\mathcal{K}}^{h}$, i.e., the node that contains the local optimum at depth $h$, is inside $\overline{\mathcal{K}}^{h}$, then we have the following inequalities
\begin{equation}
\begin{aligned}
         {f}_m(x_{h, i}) + \nu_1 \rho^{h} + 2 b_{m, h, i} 
       &\geq  \widehat{\mu}_{m, h, i} + \nu_1 \rho^{h} + b_{m, h, i} 
       \geq  \widehat{\mu}_{m, h, i_m^{p}} - b_{m, h, i_m^{p}}  \geq  \widehat{\mu}_{m, h, i_m^{*}} - b_{m, h, i_m^{p}}   \\
       &=  \overline{\mu}_{h, i_m^{*}} -  b_{m, h, i_m^{p}}
       \geq  \overline{f}(x_{h, i_m^{*}})  - b_{h, i_m^{*}}  - b_{m, h, i_m^{p}}   \geq  \overline{f}^* - 6\nu_1 \rho^{h}  - b_{h, i_m^{*}}  - b_{m, h, i_m^{p}}
\end{aligned}
\end{equation}
where the equality is because we have $\widehat{\mu}_{m, h, i_m^{*}} = \overline{\mu}_{h, i_m^{*}}$ and $  b_{m, h, i_m^{*}} =  b_{h, i_m^{*}}$ in Algorithm \ref{alg: client}. The last inequality is from Lemma \ref{lem: global_remaining_nodes}, because $(h, i_m^{*})\in \overline{\mathcal{K}}^{h}$ and thus it is uneliminated. Now we know that 
\begin{equation}
\begin{aligned}
      {f}_m^* - {f}_m(x_{h, i}) 
      &\leq   {f}_m^* - \overline{f}^* + \overline{f}^* - {f}_m(x_{h, i})
      \leq \Delta + \overline{f}^* - {f}_m(x_{h, i}) \\
      & \leq \Delta + 7 \nu_1 \rho^h + b_{m, h, i_m^{*}}  + b_{m, h, i_m^{p}} +  2 b_{m, h, i} 
      \leq 11 \nu_1 \rho^h +  \Delta
\end{aligned}
\end{equation}

\hfill $\square$

\begin{lemma}
\label{lem: optimality_lipschitz}
\textbf{\upshape (Lemma 3 in \citet{bubeck2011X})}
    For a node $\mathcal{P}_{h,i}$, define $f^*_{h,i} = \sup_{x\in \mathcal{P}_{h,i}} f(x)$ to be the maximum of the function on that region. Suppose that $f^* - f^*_{h,i} \leq c \nu_1 \rho^h$ for some $c \geq 0$, then all $x$ in $\mathcal{P}_{h,i}$ are $\max\{2c, c+1\} \nu_1 \rho^h$-optimal.
\end{lemma}

%% file: Sections/Appendix02.tex
\section{Main Proofs}
\label{app: Main_Proof}

In this section, we provide the proofs of the main theorem (Theorem \ref{thm: regret_upper_bound}) in this paper.

\textbf{Proof.} Let $E_t$ be the high probability event in Lemma \ref{lem: good_event}. Let $\mathbb{I}_{E_t}$ denote whether the event $E_t$ is true, i.e., $\mathbb{I}_{E_t} = 1$ if $E_t$ is true and 0 otherwise. We first decompose the regret into two terms
\begin{equation}
\begin{aligned}
    R(T) & =\sum_{m=1}^M \sum_{t=1}^T \left( f_m^* - f(x_{m, t})\right) =\sum_{m=1}^M \sum_{t=1}^T  \left( f_m^* - f(x_{m, t})\right)  \mathbb{I}_{E_t} + \sum_{m=1}^M \sum_{t=1}^T  \left( f_m^* - f(x_{m, t})\right) \mathbb{I}_{E_t^c} \\
    &= R(T)^{E}  + R(T)^{E^c}.
\end{aligned}
\end{equation}
For the second term, note that we can bound its expectation as follows
\begin{equation}
\begin{aligned}
\label{eqn: bound_on_Regret_Ec}
      \mathbb{E}\left[R(T)^{E^c}\right] &= \mathbb{E}\left[\sum_{m=1}^M \sum_{t=1}^T \left( f_m^* - f(x_{m, t})\right)  \mathbb{I}_{E_t^c}\right]  \leq \sum_{m=1}^M \sum_{t=1}^T \mathbb{P}\left(E_t^c\right) \leq \sum_{m=1}^M \sum_{t=1}^T (2\delta/T^6) = \frac{2M\delta}{T^5}.
\end{aligned}
\end{equation}
where the second inequality  follows from Lemma \ref{lem: good_event}. Now we bound the first term $R(T)^{E}$ in the decomposition under the event ${E}_t$. Let $H$ be a constant depth to be decided later, we know that the term $R(T)^{E}$ can be written into the following form
\begin{equation}
\begin{aligned}
\label{eqn: decomposed_R_T_E}
R(T)^E &= \sum_{m=1}^M \sum_{t=1}^T \left(   f_m^* - f_m(x_{m, t})  \right)   \mathbb{I}_{E_t} \\
&\leq \underbrace{\sum_{m=1}^M \sum_{h = 1}^H \sum_{(h,i)\in \mathcal{K}^h} \left(  f_m^* - f_m(x_{h, i}) \right ) \left \lceil \frac{\tau_{h}}{M}  \right  \rceil}_{(a)}+  \underbrace{\sum_{m=1}^M \sum_{h = 1}^H   \sum_{(h,i)\in \mathcal{K}_m^h \setminus \mathcal{K}^h} \left(  f_m^* - f_m(x_{h, i}) \right )  \tau_{h}}_{(b)}   \\
& \qquad + \underbrace{\sum_{m=1}^M \sum_{t=1}^T \sum_{h_t > H} \left(  f_m^* - f_m(x_{h_t, i_t}) \right ) }_{(c)} \\
\end{aligned}
\end{equation}

At every depth $h > 0$, for the globally un-eliminated nodes at the previous depth, i.e., for any $\mathcal{P}_{h-1,j}$ such that $(h-1, j) \in \overline{\mathcal{K}}^{h-1}$, by Lemma \ref{lem: global_remaining_nodes}, we have
\begin{equation}
    \overline{f}^* - \overline{f}(x_{h-1,j}) \leq 6 \nu_1 \rho^{h-1}.
\end{equation}
By setting $\Delta = \nu_1 \rho^{H-1}$ (to be explicitly defined later), for the locally un-eliminated nodes at the previous depth, i.e., for any $\mathcal{P}_{h-1,j}$ such that $(h-1, j) \in \overline{\mathcal{K}}_m^{h-1} \setminus \overline{\mathcal{K}}^{h-1}$, by Lemma \ref{lem: local_remaining_nodes}, we have the following inequality
    \begin{equation}
        {f}_m^* - {f}_m(x_{h-1, j}) \leq (11 \nu_1 \rho^{h-1} + \Delta) 
    \end{equation}
By Lemma \ref{lem: optimality_lipschitz} and Assumption \ref{assumption: near_optimal_similarity}, since the set ${\mathcal{K}}^h$ is created by expanding $\overline{\mathcal{K}}^{h-1}$, for the representative point $x_{h, i}$ of the node $\mathcal{P}_{h, i}$ such that $(h, i) \in \mathcal{K}^h$, we have the following upper bound on the suboptimality gap at the point $x_{h,i}$ when $h \leq H_0$.
\begin{equation}
   {f}_m^* - \overline{f}(x_{h,i}) \leq  \overline{f}^* -  \overline{f}(x_{h,i}) +  \Delta \leq (12 \nu_1 \rho^{h-1} + \Delta) \leq 13 \nu_1 \rho^{h-1} 
\end{equation}
Similarly by Lemma \ref{lem: optimality_lipschitz}, since the set ${\mathcal{K}}_m^h$ is created by expanding $\overline{\mathcal{K}}_m^{h-1}$, therefore for the representative point $x_{h, i}$ of the node $\mathcal{P}_{h, i}$ such that $(h, i) \in \mathcal{K}_m^h \setminus {\mathcal{K}}^h$, we have the following upper bound on the suboptimality gap at the point $x_{h,i}$ when $h \leq H_0$.
\begin{equation}
   {f}_m^* - {f}_m(x_{h,i}) \leq  24 \nu_1 \rho^{h-1}.
\end{equation}

\begin{itemize}
    \item In the case when $H \leq H_0$,  we know that for term (a) in Eqn. \eqref{eqn: decomposed_R_T_E}, we have 
\begin{equation}
\begin{aligned}
\label{eqn: bound_on_term_a}
(a) &\leq \sum_{h = 1}^H  \left \lceil \frac{\tau_{h}}{M}  \right \rceil \sum_{(h,i)\in \mathcal{K}^h} \sum_{m=1}^M  \left(  f_m^* - f_m(x_{h, i}) \right ) \leq  \sum_{h = 1}^H  \left \lceil \frac{\tau_{h}}{M}  \right \rceil \sum_{(h,i)\in \mathcal{K}^h} \sum_{m=1}^M  \left(  f_m^* - \overline{f}(x_{h, i}) \right )  \\
&\leq \sum_{h>0}^H 13M\nu_1 \rho^{h-1}  \max\left\{1, \frac{4c^2 \log(c_1T/\delta)}{  M\nu_1^2}  \rho^{-2h} \right\}  k |\overline{\mathcal{K}}^{h-1}| \\
&\leq \sum_{0 < h \leq h_0} 13 k CM\nu_1 \rho^{h-1}|\overline{\mathcal{K}}^{h-1}|  +  \frac{52 kc^2 C \log(c_1T/\delta)}{\nu_1 \rho^2}  \sum_{h = 1}^H  \rho^{-2h} |\overline{\mathcal{K}}^{h-1}|   \\
&\leq \sum_{0 < h \leq h_0} 13 k CM\nu_1 (\rho^{h-1})^{-(\overline{d} - 1)} +  \frac{52 kc^2 C \log(c_1T/\delta)}{\nu_1 \rho^2}  \sum_{h = 1}^H  (\rho^{h-1})^{-(\overline{d} + 1)}     \\
&\leq \sum_{0 < h \leq h_0} 13 k CM\nu_1 (\rho^{h-1})^{-(\overline{d} - 1)} +  \frac{52 kc^2 C \log(c_1T/\delta)}{\nu_1 \rho^2  (\rho^{-(\overline{d}  + 1)} - 1)}  \rho^{-H(\overline{d} +1)}    \\
\end{aligned}
\end{equation}
where $h_0 = \lfloor \frac{1}{2} \log_{\rho^{-1}} \frac{M\nu_1^2}{4c^2} \rfloor$. For the term (b), we have the following inequality
\begin{equation}
\begin{aligned}
\label{eqn: bound_on_term_b}
(b) & \leq \sum_{m=1}^M \sum_{h = 1}^H   \sum_{(h,i)\in \mathcal{K}_m^h \setminus \mathcal{K}^h} \left(  f_m^* - f_m(x_{h, i}) \right )  \tau_{h}
\leq   \sum_{m=1}^M \sum_{h = 1}^H   \sum_{(h,i)\in \mathcal{K}_m^h \setminus \mathcal{K}^h} \frac{48c^2 \log(c_1T/\delta)}{  \nu_1 \rho^2}  \rho^{-(h-1)}  \\
& \leq  \sum_{m=1}^M \sum_{h = 1}^H \frac{48c^2 \log(c_1T/\delta)}{  \nu_1 \rho^2}  \rho^{-(h-1)} k |\overline{\mathcal{K}}_m^{h-1}|  \leq \frac{48 kc^2 C  \log(c_1T/\delta)}{  \nu_1 \rho^2}  \sum_{h = 0}^{H-1}   M \rho^{-h(d_{\max} +1)}  \\
&= \frac{48kc^2 C  \log(c_1T/\delta)}{  \nu_1 \rho^2 (\rho^{-(d_{\max}  + 1)} - 1)} M \rho^{-H(d_{\max} +1)} 
\end{aligned}
\end{equation}
For term (c), it could be bounded by
\begin{equation}
\begin{aligned}
\label{eqn: bound_on_term_c}
(c) & \leq \sum_{t=1}^T  24 M \nu_1 \rho^{H} \leq 24 M \nu_1 \rho^{H} T
\end{aligned}
\end{equation}
Therefore if we combine the bounds on the three terms (a), (b), and (c), in Eqns. \eqref{eqn: bound_on_term_a}, \eqref{eqn: bound_on_term_b}, \eqref{eqn: bound_on_term_c}, we have the following inequality
\begin{equation}
\begin{aligned}
& R(T)^E  \leq (a) + (b) + (c) \\
&\leq C_0 +  \frac{52 kc^2 C \log(c_1T/\delta)}{\nu_1 \rho^2  (\rho^{-(\overline{d}  + 1)} - 1)}  \rho^{-H(\overline{d} +1)}  + \frac{48kc^2 C  \log(c_1T/\delta)}{  \nu_1 \rho^2 (\rho^{-(d_{\max} + 1)} - 1)} M \rho^{-H(d_{\max} +1)} + 24 \nu_1 \rho^{H} MT \\
&\leq C_0 +  2 \max \left\{\frac{52 kc^2 C \log(c_1T/\delta)}{\nu_1 \rho^2  (\rho^{-(\overline{d}  + 1)} - 1)}  \rho^{-H(\overline{d} +1)}, \frac{48kc^2 C  \log(c_1T/\delta)}{  \nu_1 \rho^2 (\rho^{-(d_{\max} + 1)} - 1)} M \rho^{-H(d_{\max} +1)} \right \}+ 24 \nu_1 \rho^{H} MT \\
&\leq C_0 +  C_1 \max \left \{M^\frac{\overline{d}  + 1}{\overline{d}  + 2} T^\frac{\overline{d}  + 1}{\overline{d}  + 2} (\log(MT))^\frac{\overline{d}  + 1}{\overline{d}  + 2},  M T^\frac{{d}_{\max}  + 1}{{d}_{\max}  + 2}(\log(M T))^\frac{{d}_{\max}  + 1}{{d}_{\max}  + 2} \right\}
\end{aligned}
\end{equation}
where $C_0$ is a constant, $C_1 = (2+2\log c_1) \max \left\{\left(\frac{52  kc^2 C (24\nu_1)^{(\overline{d}  + 1)}}{\nu_1 \rho^2  (\rho^{-(\overline{d}  + 1)} - 1)}\right)^{\frac{1}{\overline{d}  + 2}},  \left(\frac{48  kc^2 C (24\nu_1)^{( {d}_{\max}  + 1)}}{\nu_1 \rho^2  (\rho^{-( {d}_{\max}  + 1)} - 1)}\right)^{\frac{1}{d_{\max}  + 2}} \right\}$ and the last inequality is by balancing the size of the dominating terms using $H$. Now combining all the above bounds on $R(T)^{E}$ and $R(T)^{E^c}$, we know that the regret is of order $\widetilde{\mathcal{O}} \left( M^\frac{\overline{d}  + 1}{\overline{d}  + 2} T^\frac{\overline{d}  + 1}{\overline{d}  + 2} + M T^\frac{{d}_{\max}  + 1}{{d}_{\max}  + 2}\right)$.

\item In the case when $H \geq H_0$, it means that the federated learning process terminated before even reaching $H$, then the clients optimize their local objectives separately. Therefore, we know that for term (a) in Eqn. \eqref{eqn: decomposed_R_T_E}, we have 
\begin{equation}
\begin{aligned}
(a) &\leq \sum_{h = 1}^H  \left \lceil \frac{\tau_{h}}{M}  \right \rceil \sum_{(h,i)\in \mathcal{K}^h} \sum_{m=1}^M  \left(  f_m^* - f_m(x_{h, i}) \right ) \leq  \sum_{h = 1}^H  \left \lceil \frac{\tau_{h}}{M}  \right \rceil \sum_{(h,i)\in \mathcal{K}^h} \sum_{m=1}^M  \left(  f_m^* - \overline{f}(x_{h, i}) \right )  \\
&\leq \sum_{h>0}^{H_0} 13M\nu_1 \rho^{h-1}  \max\left\{1, \frac{4c^2 \log(c_1T/\delta)}{  M\nu_1^2}  \rho^{-2h} \right\}  k |\overline{\mathcal{K}}^{h-1}| \\
&\leq \sum_{0 < h \leq h_0} 13 k CM\nu_1 \rho^{h-1}|\overline{\mathcal{K}}^{h-1}|  +  \frac{52 kc^2 C \log(c_1T/\delta)}{\nu_1 \rho^2}  \sum_{h = 1}^{H_0}  \rho^{-2h} |\overline{\mathcal{K}}^{h-1}|   \\
&\leq \sum_{0 < h \leq h_0} 13 k CM\nu_1 (\rho^{h-1})^{-(\overline{d} - 1)} +  \frac{52 kc^2 C \log(c_1T/\delta)}{\nu_1 \rho^2}  \sum_{h = 1}^{H_0}  (\rho^{h-1})^{-(\overline{d} + 1)}     \\
&\leq \sum_{0 < h \leq h_0} 13 k CM\nu_1 (\rho^{h-1})^{-(\overline{d} - 1)} +  \frac{52 kc^2 C \log(c_1T/\delta)}{\nu_1 \rho^2  (\rho^{-(\overline{d}  + 1)} - 1)}  \rho^{-H_0(\overline{d} +1)}    \\
\end{aligned}
\end{equation}
where $h_0 = \lfloor \frac{1}{2} \log_{\rho^{-1}} \frac{M\nu_1^2}{4c^2} \rfloor$. For the term (b), we have the following inequality
\begin{equation}
\begin{aligned}
(b) & \leq \sum_{m=1}^M \sum_{h = 1}^H   \sum_{(h,i)\in \mathcal{K}_m^h \setminus \mathcal{K}^h} \left(  f_m^* - f_m(x_{h, i}) \right )  \tau_{h}
\leq   \sum_{m=1}^M \sum_{h = 1}^H   \sum_{(h,i)\in \mathcal{K}_m^h \setminus \mathcal{K}^h} \frac{48c^2 \log(c_1T/\delta)}{  \nu_1 \rho^2}  \rho^{-(h-1)}  \\
& \leq  \sum_{m=1}^M \sum_{h = 1}^H \frac{48c^2 \log(c_1T/\delta)}{  \nu_1 \rho^2}  \rho^{-(h-1)} k |\overline{\mathcal{K}}_m^{h-1}|  \leq \frac{48 kc^2 C  \log(c_1T/\delta)}{  \nu_1 \rho^2}  \sum_{h = 0}^{H-1}   M \rho^{-h(d_{\max} +1)}  \\
&= \frac{48kc^2 C  \log(c_1T/\delta)}{  \nu_1 \rho^2 (\rho^{-(d_{\max}  + 1)} - 1)} M \rho^{-H(d_{\max} +1)} 
\end{aligned}
\end{equation}
where the second inequality is because when $h \in [1, H_0]$, we have ${f}_m^* - {f}_m(x_{h,i}) \leq  24 \nu_1 \rho^{h-1}$. When $h > H_0$, it means that the clients start learning separately and thus we have ${f}_m^* - {f}_m(x_{h, i})  \leq 12 \nu_1 \rho^{h-1}$ by Lemma \ref{lem: local_remaining_nodes}. Therefore in the worst case, ${f}_m^* - {f}_m(x_{h,i}) \leq  24 \nu_1 \rho^{h-1}$.  For term (c), it could be bounded by
\begin{equation}
\begin{aligned}
(c) & \leq \sum_{t=1}^T  12 M \nu_1 \rho^{H} \leq 12 M \nu_1 \rho^{H} T
\end{aligned}
\end{equation}
Therefore if we combine the bounds on the three terms (a), (b), and (c), we have the following inequality
\begin{equation}
\begin{aligned}
& R(T)^E  \leq (a) + (b) + (c) \\
&\leq C_0 +  \frac{52 kc^2 C \log(c_1T/\delta)}{\nu_1 \rho^2  (\rho^{-(\overline{d}  + 1)} - 1)}  \rho^{-H_0(\overline{d} +1)}  + \frac{48kc^2 C  \log(c_1T/\delta)}{  \nu_1 \rho^2 (\rho^{-(d_{\max} + 1)} - 1)} M \rho^{-H(d_{\max} +1)} + 12 \nu_1 \rho^{H} MT \\
&\leq C_0 +  \frac{52 kc^2 C \log(c_1T/\delta)}{\nu_1 \rho^2  (\rho^{-(\overline{d}  + 1)} - 1)}  \rho^{-H(\overline{d} +1)}  + \frac{48kc^2 C  \log(c_1T/\delta)}{  \nu_1 \rho^2 (\rho^{-(d_{\max} + 1)} - 1)} M \rho^{-H(d_{\max} +1)} + 12 \nu_1 \rho^{H} MT \\
&\leq C_0 +  2 \max \left\{\frac{52 kc^2 C \log(c_1T/\delta)}{\nu_1 \rho^2  (\rho^{-(\overline{d}  + 1)} - 1)}  \rho^{-H(\overline{d} +1)}, \frac{48kc^2 C  \log(c_1T/\delta)}{  \nu_1 \rho^2 (\rho^{-(d_{\max} + 1)} - 1)} M \rho^{-H(d_{\max} +1)} \right \}+ 12 \nu_1 \rho^{H} MT \\
&\leq C_0 +  C_1 \max \left \{M^\frac{\overline{d}  + 1}{\overline{d}  + 2} T^\frac{\overline{d}  + 1}{\overline{d}  + 2} (\log(MT))^\frac{\overline{d}  + 1}{\overline{d}  + 2},  M T^\frac{{d}_{\max}  + 1}{{d}_{\max}  + 2}(\log(M T))^\frac{{d}_{\max}  + 1}{{d}_{\max}  + 2} \right\}
\end{aligned}
\end{equation}
where $C_0$ is a constant, $C_1 = (2+2\log c_1) \max \left\{\left(\frac{52  kc^2 C (12\nu_1)^{(\overline{d}  + 1)}}{\nu_1 \rho^2  (\rho^{-(\overline{d}  + 1)} - 1)}\right)^{\frac{1}{\overline{d}  + 2}},  \left(\frac{48  kc^2 C (12\nu_1)^{( {d}_{\max}  + 1)}}{\nu_1 \rho^2  (\rho^{-( {d}_{\max}  + 1)} - 1)}\right)^{\frac{1}{d_{\max}  + 2}} \right\}$ and the last inequality is by balancing the size of the dominating terms using $H$. Now combining all the above bounds on $R(T)^{E}$ and $R(T)^{E^c}$, we know that the regret is of order $\widetilde{\mathcal{O}} \left( M^\frac{\overline{d}  + 1}{\overline{d}  + 2} T^\frac{\overline{d}  + 1}{\overline{d}  + 2} + M T^\frac{{d}_{\max}  + 1}{{d}_{\max}  + 2}\right)$.
\end{itemize}

\hfill $\square$

\subsection{Proof of Corollary \ref{corollary: regret_upper_bound}}

When we assume Assumption \ref{assumption: near_optimal_similarity}, we basically assume that near-optimal nodes in $\overline{f}$ are also near-optimal in the local objectives. Without loss of generality, we assume that $\omega = 1$ in Assumption \ref{assumption: near_optimal_similarity}. If $\omega \neq 1$, we only have to change a few constants in the proof.

For term $(a)$ in Eqn. \eqref{eqn: decomposed_R_T_E}, note that every $6\nu_1\rho^h$-near-optimal node for $\overline{f}$ is $\Delta + 6\nu_1 \rho^h \leq 12\nu_1\rho^h$-near-optimal in every $f_m$, that means $\overline{\mathcal{K}}^{h-1} \subseteq \overline{\mathcal{K}}_m^{h-1}, \forall m \in [M], \forall h \leq H$. Therefore, we could bound term $(a)$ as
\begin{equation}
\begin{aligned}
(a)  \leq \sum_{0 < h \leq h_0} 13 k CM\nu_1 (\rho^{h-1})^{-(d_{\min}- 1)} +  \frac{52 kc^2 C \log(c_1T/\delta)}{\nu_1 \rho^2  (\rho^{-(d_{\min}  + 1)} - 1)}  \rho^{-H_0(d_{\min} +1)}    
\end{aligned}
\end{equation}
whereas for term $(b)$ in Eqn. \eqref{eqn: decomposed_R_T_E}, since $\mathcal{N}_{f_m}(12\nu\rho^h, \nu\rho^h) \setminus \mathcal{N}_{\overline{f}}(6\nu\rho^h, \rho^h) \leq C_0 \rho^{-d_{\text{new}}h}, \forall m \in [M]$, we have the following bound
\begin{equation}
\begin{aligned}
(b) &\leq \frac{48kc^2 C_0  \log(c_1T/\delta)}{  \nu_1 \rho^2 (\rho^{-(d_{\text{new}}  + 1)} - 1)} M \rho^{-H(d_{\text{new}} +1)} 
\end{aligned}
\end{equation}
If we combine the bounds on the three terms (a), (b), and (c), we have the following inequality
\begin{equation}
\begin{aligned}
& R(T)^E  \leq (a) + (b) + (c) \\
&\leq C_0' +  \frac{52 kc^2 C \log(c_1T/\delta)}{\nu_1 \rho^2  (\rho^{-(d_{\min}  + 1)} - 1)}  \rho^{-H_0(d_{\min} +1)} +  \frac{48kc^2 C_0  \log(c_1T/\delta)}{  \nu_1 \rho^2 (\rho^{-(d_{\text{new}}  + 1)} - 1)} M \rho^{-H(d_{\text{new}} +1)}    + 24 \nu_1 \rho^{H} MT \\
&\leq C_0' +  2 \max \left\{\frac{52 kc^2 C \log(c_1T/\delta)}{\nu_1 \rho^2  (\rho^{-(d_{\min} + 1)} - 1)}  \rho^{-H(d_{\min}+1)}, \frac{48kc^2 C  \log(c_1T/\delta)}{  \nu_1 \rho^2 (\rho^{-(d_{\text{new}} + 1)} - 1)} M \rho^{-H(d_{\text{new}} +1)} \right \}+ 24 \nu_1 \rho^{H} MT \\
&\leq C_0' +  C_1' \max \left \{M^\frac{d_{\min} + 1}{d_{\min} + 2} T^\frac{d_{\min} + 1}{d_{\min} + 2} (\log(MT))^\frac{d_{\min} + 1}{d_{\min} + 2},  M T^\frac{d_{\text{new}}  + 1}{d_{\text{new}}  + 2}(\log(M T))^\frac{d_{\text{new}}  + 1}{d_{\text{new}}  + 2} \right\}
\end{aligned}
\end{equation}
where $C_0' > 0, C_1' > 0$ is another set of constants. Therefore the final regret for the \texttt{PF-PNE} algorithm is bounded by $\widetilde{\mathcal{O}} \left( M^\frac{d_{\min}  + 1}{d_{\min}   + 2} T^\frac{d_{\min}   + 1}{d_{\min} + 2} + M T^\frac{d_{\text{new}}  + 1}{d_{\text{new}}  + 2}\right)$.

%% file: Sections/Appendix03.tex
\section{Experimental Details}
\label{app: experimental_details}

In this section, we provide all the details related to the algorithms, datasets, and hyper-parameters in Section \ref{sec: experiments}. We also provide more federated $\mathcal{X}$-armed bandit experiments.

\subsection{Algorithms and Hyper-parameters}

For the implementation of hierarchical partitioning and centralized $\mathcal{X}$-armed bandit algorithms, we have used the publicly available open-source package PyXAB by \citet{Li2023PyXAB}. We list the algorithms used in our experiments and the hyper-parameter settings of these algorithms.

\begin{list1}
    \item \textbf{\texttt{HCT}.}
The \texttt{HCT} algorithm is a (single-client) $\mathcal{X}$-armed bandit algorithm proposed by \citet{azar2014online}. We have used the publicly-available implementation by \citet{Li2023PyXAB} at the link \url{https://github.com/WilliamLwj/PyXAB}

\item \textbf{\texttt{Fed1-UCB}. }
 The \texttt{Fed1-UCB} algorithm is a multi-armed bandit algorithm proposed by \citet{shi2021federateda}. We have followed \citet{Li2022Federated} and generate 20 arms on each dimension randomly for each trial of the algorithm for 1-D and 2-D objective functions. For other high-dimensional functions, we have randomly generated 1000 arms for Fed1-UCB. The hyper-parameters are set to be the same as the original paper and their codebase.

\item \textbf{\texttt{FN-UCB}.}
The \texttt{FN-UCB} algorithm is a neural bandit  algorithm proposed by \citet{dai2023federated}. We have used the public implementation  \citet{dai2023federated} at the link \url{https://github.com/daizhongxiang/Federated-Neural-Bandits} with the default hyperparamter choices. Similar to \texttt{FN-UCB}, we have generated 20 arms on each dimension randomly for each trial of the algorithm for 1-D and 2-D objective functions. For other high-dimensional functions, we have randomly generated 1000 arms.

\item \textbf{\texttt{Fed-PNE}.}
We have followed \citet{Li2022Federated} and used their parameter settings for the \texttt{Fed-PNE} algorithm.

\begin{itemize}
    \item The smoothness parameters $\nu_1$ and $\rho$ are set to be $\nu_1 = 1$ and $\rho = 0.5$.
    \item The confidence parameters $c$ and $c_1$ are set to be $c = 0.1$ and $c_1 = 1$. 
\end{itemize}

 Notably, the performance of \texttt{Fed-PNE} is originally measured by the global regret, i.e., the regret on the average of all local objective. However, in this paper we measure the performance on the local objectives. We believe this is the main reason why \texttt{Fed-PNE} performs non-ideally in our experiments.

\item \textbf{\texttt{PF-PNE}.}
Since \texttt{PF-PNE} can be viewed as an ``upgraded" version of \texttt{Fed-PNE}, we have used the same hyper-parameter setting as the \texttt{Fed-PNE}, i.e., $\nu_1 = 1, \rho = 0.5, c = 0.1$ and $c_1 = 1$. For the additional hyper-parameter $\Delta$, we have set it to be $\Delta = 0.01$ in all the experiments. Tuning these hyper-parameters will not affect the final result too much.
\end{list1}

\subsection{Objective Functions and Dataset}

\textbf{Synthetic Functions}. Garland, DoubleSine, Himmelblau, and Rastrigin are synthetic functions that are used very frequently in the experiments of $\mathcal{X}$-armed bandit algorithms because of their large number of local optimums and their extreme unsmoothness, which appeared in works such as \citet{azar2014online, Grill2015Blackbox, shang2019general, bartlett2019simple, li2021optimumstatistical}. Garland and DoubleSine are defined on the domain $[0, 1]$, Himmelblau is defined on $[-5, 5]$, while Rastrigin can be defined on $[-1, 1]^k$ where $k$ is an arbitrarily large integer. We have normalized these functions so that their values are between [0, 1] to fulfill the requirements in the analysis. The local objectives are the shifted versions of the original objectives, with a random shift on each dimension. Random noise is added to the function evaluations.

\textbf{Landmine Dataset}.
The landmine dataset contains multiple landmine fields with features from radar images. We have followed \citet{dai2020federated} and split the dataset into equal-sized training set and testing set. Each client randomly chooses one landmine field and optimize one SVM machine to detect the landmines in the particular field. The local objectives are the AUC-ROC scores on one landmine objective. The original dataset can be downloaded from \url{http://www.ee.duke.edu/~lcarin/ LandmineData.zip}

\subsection{Additional Experiments}

\begin{figure}
\centering
\hspace*{-1em}
\subfigure[\footnotesize Doublesine (1D)]{
  \centering
  \includegraphics[width=0.44\linewidth]{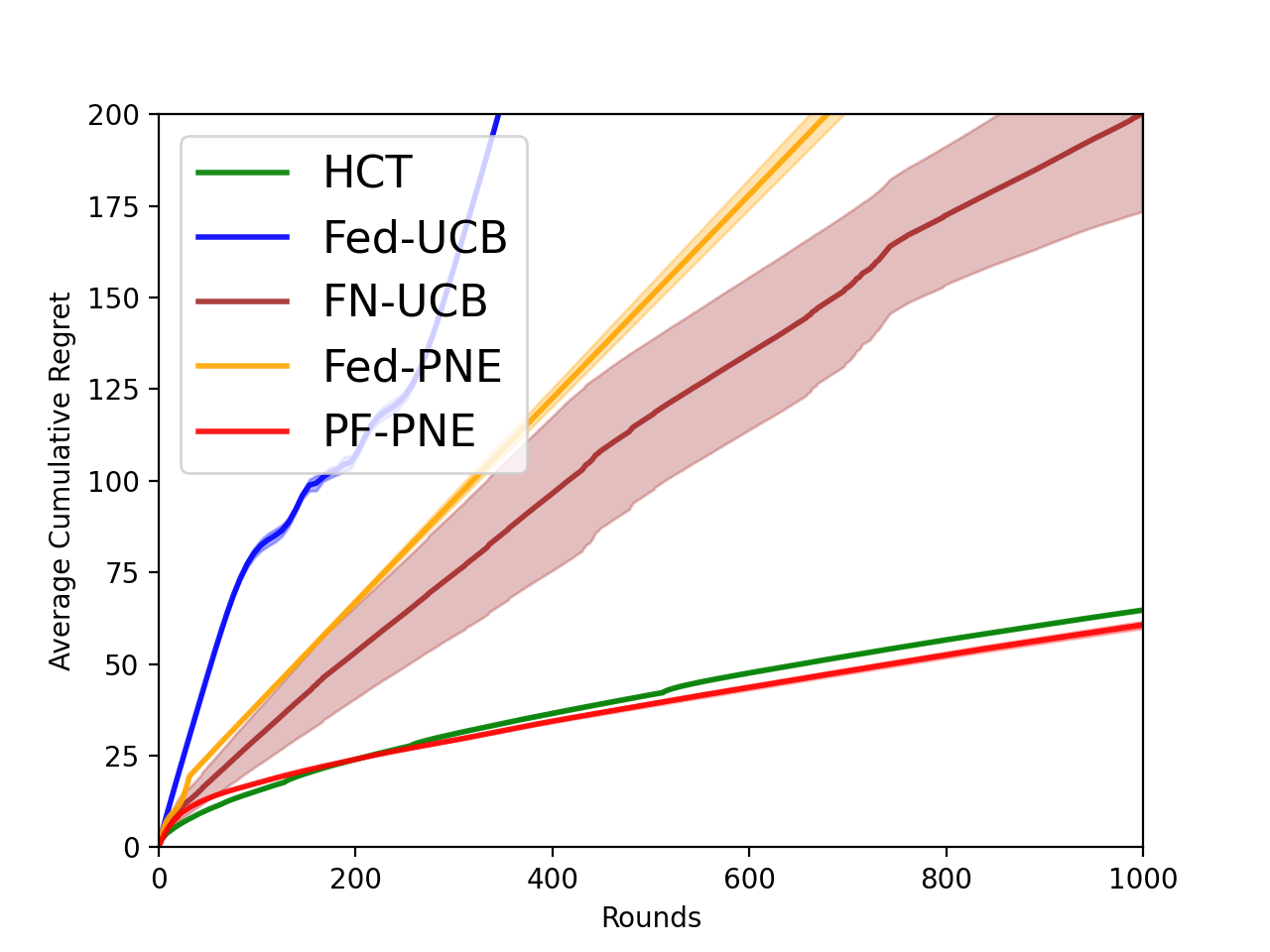}
  \label{fig: Doublesine}
}\hspace*{-2em}%
\subfigure[\footnotesize Ackley (2D)]{
  \centering
  \includegraphics[width=0.44\linewidth]{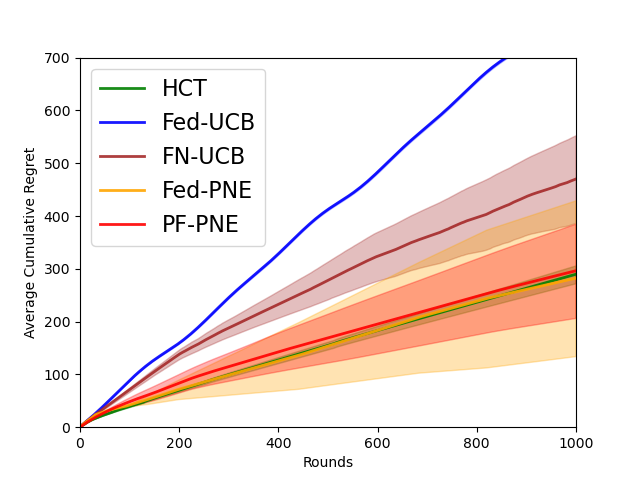}
  \label{fig: Ackley}
}\hspace*{-2em}%
\vspace{-10pt}
\caption{Cumulative regret of different algorithms on the synthetic functions. Unlimited communications are allowed for centralized algorithms.}
\vspace{-10pt}
\label{fig: app_experiments}
\end{figure}

We have conducted additional experiments on two more synthetic objectives Doublesine (1D) and Ackley (2D). Similarly, we add random shifts to the each dimension of the original objectives to produce the local objectives of each client. The experimental results are similar to what we present in the main paper. \texttt{PF-PNE} performs slightly better than \texttt{HCT} and much better than \texttt{Fed-PNE} on Doublesine. On the other hand, \texttt{PF-PNE} performs similarly as \texttt{HCT} and \texttt{Fed-PNE} on Ackley. Both results are aligned with our theoretical analysis.

\subsection{Communication Cost}

\begin{figure}[ht]
\centering
\hspace*{-1em}
\subfigure[\footnotesize Garland (1D)]{
  \centering
  \includegraphics[width=0.27\linewidth]{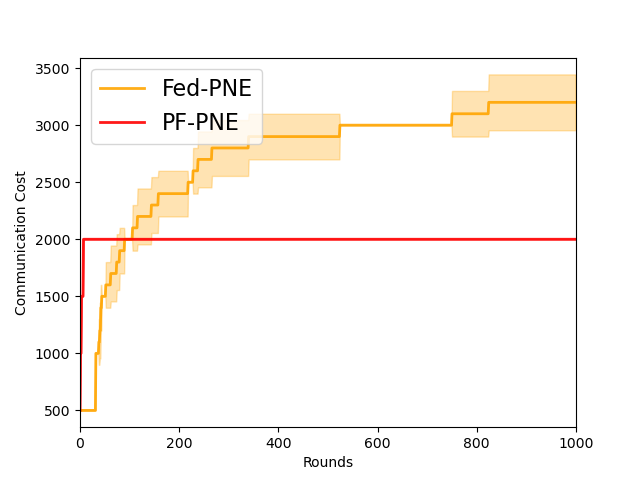}
  \label{fig: Garland_comm_cost}
}\hspace*{-1em}%
% \subfigure[\footnotesize {DoubleSine}]{
%   \centering
%   \includegraphics[width=0.28\linewidth]{figs_pn/doublesine.png}
%     \label{fig: DoubleSine}
%}\hspace*{-1.5em}%
\subfigure[\footnotesize Himmelblau (2D)]{
  \centering
  \includegraphics[width=0.27\linewidth]{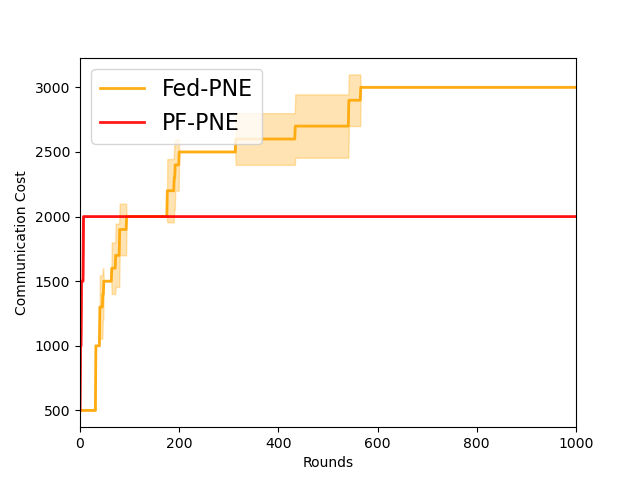}
  \label{fig: Himmelblau_comm_cost}
}\hspace*{-1em}%
\subfigure[\footnotesize Rastrigin (10D)]{
  \centering
  \includegraphics[width=0.27\linewidth]{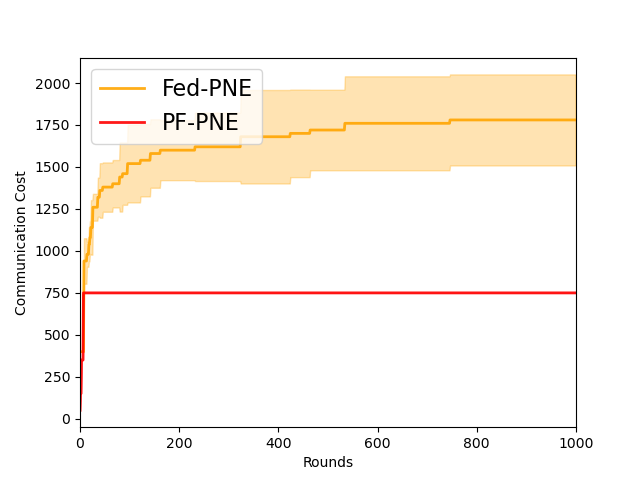}
  \label{fig: Rastrigin_comm_cost}
}\hspace*{-1em}%
\subfigure[\footnotesize Ackley (2D)]{
  \centering
  \includegraphics[width=0.27\linewidth]{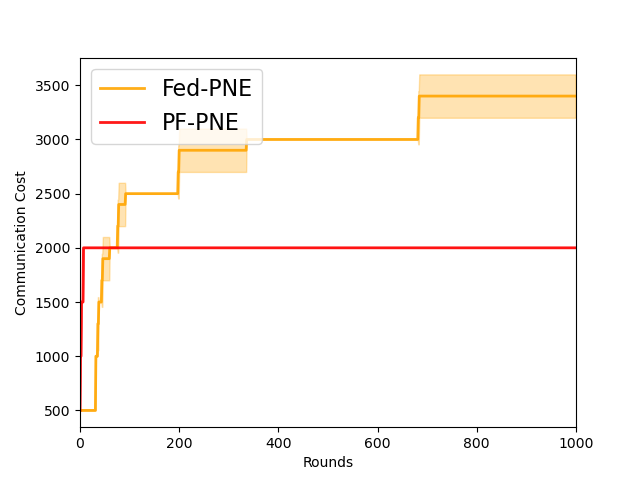}
  \label{fig: Ackley_comm_cost}
}
\caption{Communication cost comparison between \texttt{Fed-PNE} and \texttt{PF-PNE}}
\label{fig: experiments_comm_cost}
\end{figure}

We provide the communication cost comparison between \texttt{Fed-PNE} and \texttt{PF-PNE} on the synthetic objectives, as shown in Figure \ref{fig: experiments_comm_cost}. As can be observed, the communication cost of \texttt{Fed-PNE} keeps increasing. However, the communication cost of \texttt{PF-PNE} stops to increase after a certain point in the learning process, proving the correctness of our theory.

